%% file: 2025-ral-gaussianProcessAero.tex
\documentclass[letterpaper, 10 pt, conference]{ieeeconf}
\IEEEoverridecommandlockouts  % This command is only needed if you want to use the \thanks command

\input{preamble/common}
\newcommenter{leo}{Firebrick1}
\newcommenter{sgrace}{Purple2}
\newcommenter{arozman}{Orange2}

\title{\LARGE \bf
  Gradient-Enhanced Partitioned Gaussian Processes for Real-Time Quadrotor Dynamics Modeling
}

\author{Xinhuan Sang, Adam Rozman, Sheryl Grace, and Roberto Tron% <-this % stops a space
  \thanks{The authors are with the Department of Mechanical Engineering,
    Boston University, 110 Cummington Mall, MA 02215, United States
    {\tt\small \{leosang,arozman,sgrace,tron\}@bu.edu}}%
  \thanks{Support for this work was provided by NASA ULI grant 9500314748 ``Safe, Low-Noise Operation of UAM in Urban Canyons via Integration of Gust Outcomes and Trim Optimization''}
  \thanks{This work has been submitted to the IEEE for possible publication. Copyright may be transferred without notice, after which this version may no longer be accessible.}
}

\begin{document}

\maketitle
\thispagestyle{empty}
\pagestyle{empty}

%%%%%%%%%%%%%%%%%%%%%%%%%%%%%%%%%%%%%%%%%%%%%%%%%%%%%%%%%%%%%%%%%%%%%%%%%%%%%%%%
\begin{abstract}
  We present a quadrotor dynamics Gaussian Process (GP) with gradient information that achieves real-time inference via state-space partitioning and approximation, and that includes aerodynamic effects using data from mid-fidelity potential flow simulations.
  While traditional GP-based approaches provide reliable Bayesian predictions with uncertainty quantification, they are computationally expensive and thus unsuitable for real-time simulations. To address this challenge, we integrate gradient information to improve accuracy and introduce a novel partitioning and approximation strategy to reduce online computational cost. In particular, for the latter, we associate a local GP with each non-overlapping region; by splitting the training data into local \emph{near} and \emph{far} subsets, and by using Schur complements, we show that a large part of the matrix inversions required for inference can be performed offline, enabling real-time inference at frequencies above \unit[$30$]{Hz} on standard desktop hardware.
  To generate a training dataset that captures aerodynamic effects, such as rotor-rotor interactions and apparent wind direction, we use the CHARM code, which is a mid-fidelity aerodynamic solver. It is applied to the SUI Endurance quadrotor to predict force and torque, along with noise at three specified locations. The derivative information is obtained via finite differences.
  Experimental results demonstrate that the proposed partitioned GP with gradient conditioning achieves higher accuracy than standard partitioned GPs without gradient information, while greatly reducing computational time. This framework provides an efficient foundation for real-time aerodynamic prediction and control algorithms in complex and unsteady environments.
\end{abstract}

%%%%%%%%%%%%%%%%%%%%%%%%%%%%%%%%%%%%%%%%%%%%%%%%%%%%%%%%%%%%%%%%%%%%%%%%%%%%%%%%
\section{Introduction}

Unmanned aerial vehicles (UAVs) with fixed rotors are becoming increasingly integrated into daily life, from wall inspection and window cleaning for buildings to routine delivery tasks \cite{khan2018drones}. During such operations, UAVs inevitably encounter complex and rapidly varying aerodynamic environments (e.g., wind gusts). However, existing UAV dynamics simulators often do not accurately reproduce these conditions, due to the difficulty of building analytical models and collecting data from actual systems in application-relevant conditions; this represents a bottleneck for the design and evaluation of effective control algorithms in such environments.

Data-driven machine learning has been recognized as a promising approach to address this issue. Significant progress has been made in this direction. For example, deep learning has been widely applied to integrate flight data \cite{chee2022knode,shi2019neural}; however, due to the opaque nature of neural networks, these models typically lack the ability to quantify the reliability of their predictions. By contrast, Gaussian Process (GP) models provide Bayesian, closed-form predictive distributions, offering both accurate estimates and associated uncertainty quantification \cite{Rasmussen2006Gaussian}. Nonetheless, GP-based methods are often limited by their high computational complexity (which is more than $O(n^2)$ in the number of training points \cite{koller2018learning}).

To mitigate this challenge, some studies combine simple analytical models to capture the dominant dynamics with GP corrections, thus reducing the number of training points required for similar accuracy (e.g., Hewing et al. 2018 \cite{hewing2018cautious}). Other works employ sparse (e.g., Kulathunga et al. 2024 \cite{kulathunga2024sparse}; Leibfried et al. 2020 \cite{leibfried2020tutorial}) or partitioned (e.g., Nguyen-Tuong et al. 2008 \cite{nguyen2008local}; Park et al. 2016 \cite{vlastos2021partitioned}; Lee et al. 2023 \cite{lee2023partitioned}) GP variants that restrict computations to training samples near the query point. The sparse approach greatly reduces offline and online computation time via low-rank approximations but sacrifices prediction fidelity. Partitioned GPs (local GPs) are another approach to reducing the computational cost for GP by dividing the dataset or input space into multiple subregions/blocks and training a separate GP for each subregion/block. This approach better captures local variation, but introduces region-to-region boundary discontinuities and artifacts at region edges.

The training data of interest for the current application are forces and torques associated with a UAV system, as well as noise from three predefined azimuth-distance combinations. These data could be obtained through either experiment or physics-based simulation. In this work, simulations are used. As a robust database requires results for a large number of inflow conditions and UAV settings, the selection of an aerodynamic solver must weigh fidelity and computational time. For instance, computational fluid dynamics simulations solving the full governing differential equations for the unsteady fluid flow and interactions of a quadrotor can require $\mathcal{O}(10^6)$ CPU hours for a single case \cite{thai2022quad} which is computationally prohibitive to use for more than a few cases. On the other end of the spectrum, Blade Element Moment Theory (BEMT) produces a solution based completely on a simplified model of the local flow conditions across a propeller and airfoil  aerodynamic interpolated from an airfoil. This method provides a solution in seconds but cannot account for complex inflow and interaction between multiple propellers. % and uses simplified inflow models that do not often reflect the actual situation well.
% When one is interested in the averaged force, torque, and a noise value associated only with tones, such fidelity is not needed.
Therefore, a mid-fidelity solver has been selected. CHARM, a comprehensive rotorcraft analysis tool\cite{CHARM}, applies a wake model and a vortex lattice lifting line method that still relies on airfoil tables to calculate local propeller sectional forces but tracks the wake interactions of multiple bodies and computes the local flow field more realistically. With this computational tool, a single quadrotor operating point can be calculated in a minute. Once the performance is computed, the noise is calculated using PSU-WOPWOP, which will be discussed in more detail in \Cref{sec:aero_modeling}.

Our goal is to build a mid-fidelity, real-time surrogate of quadrotor forces, torques, and noise that is accurate enough to stress-test control algorithms in complex flow conditions, yet fast enough to run at \unit[$30$]{Hz} on commodity hardware.

In this study, we propose a partitioned Gaussian Process framework for mid-fidelity, real-time quadrotor dynamics prediction. As in previous work \cite{hewing2018cautious}, we employ the hybrid approach that supplements a GP with a simplified analytical model \cite{mellinger:ICRA11}.
We subtract the simplified model’s contribution from the CHARM-supplied results.
We apply numerical differentiation of the simulations above to extract derivative information, enabling a GP with gradient conditioning to learn the residual dynamics. To further reduce online computational costs, the dataset is pre-partitioned into subsets. For each subset, we apply the Schur complement to distinguish between near-field and far-field contributions and precompute the corresponding matrices.

Overall, our partitioned structure with gradient information\begin{enumerate}
\item significantly reduces the real-time computational burden of GPs with a large dataset,
\item forms the base for building a high-accuracy quadrotor dynamics simulator for the development of control algorithms,
\item has the potential to produce high-fidelity dynamics and noise predictions when given high-fidelity simulated data, and
\item can be easily applied to similar machine learning problems with a large dataset and gradient information.
\end{enumerate}

\section{Preliminary Concepts}
\subsection{Schur Complements}
\label{sec:SchurComplements}

Schur complements arise in the computation of conditional Gaussian distributions and in partial solutions to systems of linear equations.
Given a partitioned matrix $M$ of the form:
\begin{equation}
  M =
  \begin{bmatrix}
    A & B \\
    C & D
  \end{bmatrix},
\end{equation}

The Schur complement of the block $D$ of the matrix $M$ is defined by:
\begin{equation}
  M | D = A - B D\inverse C.
\end{equation}

And the Schur complement inversion of matrix $M$ is defined as:
\begin{small}\begin{multline}
\label{eq:SchurInv}
    M\inverse=\\
    \begin{bmatrix}
        (M | D)\inverse & -(M | D)\inverse BD\inverse\\
        -D\inverse C(M | D)\inverse& D\inverse+D\inverse C(M | D)\inverse BD\inverse
    \end{bmatrix}.
\end{multline}
\end{small}

\subsection{Gaussian Processes}
\label{sec:GPintro}
Gaussian Processes (GPs) are a generative machine learning model that is typically used for regression from data \cite{Rasmussen2006Gaussian}. A Gaussian process is a \emph{random process} $f(x):\mathbb{R}^{d}\to\mathbb{R}$ defined by a \emph{mean function} $\mu(x): \mathbb{R}^{d}\rightarrow\mathbb{R}$ and a \emph{covariance kernel} $k(x, x'):\mathbb{R}^{d}\times\mathbb{R}^{d}\rightarrow\mathbb{R}$:
\begin{equation}
  f(x) \sim \cN\bigl(\mu,k\bigr).
\end{equation}

The defining characteristic is that, given a set of $n_d$ data points $X_d=\operatorname{stack}(\{x_i\}_{i=1}^{n_d})$ (we use $\operatorname{stack}(\cdot)$ to denote vertical concatenation of vectors), the evaluations of the process at those points $f(X_d)\in\mathbb{R}^{n_d}$ follows a Gaussian distribution:
\begin{equation}\label{eq:distribution fd}
  f(X_d) \sim \cN(\mu_d,K_{d,d}),
\end{equation}
with mean $\mu_d\in\mathbb{R}^{n_d}$ and covariance $K_{d,d}\in\mathbb{R}^{n_d\times n_d}$ given by
\begin{subequations}
  \begin{align}
    \mu_d&=\mu(X_d),\\
    K_{d,d}&=k(X_d,X_d).
  \end{align}
\end{subequations}

In the remainder of the paper, we make the common assumption that the mean function is zero, $\mu(x)=0$, and we use a Radial Basis Function (RBF) kernel
\begin{equation}\label{eq:kernel 00}
  k(x,x') = \sigma \exp\bigl(-\tfrac{\norm{x - x'}^2}{2l^2}\bigr),\\
\end{equation}

which has two hyper parameters: signal variance $\sigma$ and radius scale $l$. The signal variance $\sigma$ adjusts the amplitude of the covariance in the GP. The radius scale $l$ adjusts the rate of diminishing as two states become distant, which, on a broader scale, controls how quickly the GP output can change with respect to input change.

\paragraph{Inference from training data}
Given a dataset $D=\{X_d, Y_d\}$ where $Y_d=\operatorname{stack}(\{y_i\}_{i=1}^{n_d})\in\real{n_d}$ is a realization (measurement) of $f(X_d)$,  and \emph{query points} $X_q=\operatorname{stack}(\{ x_{q, i}\}^{n_q}_{i=1})$, $x_{q, i}\in\mathbb{R}^{d}$, the definition of a GP implies that the conditional distribution of $f(X_q)$ given $Y_d$ is a Gaussian distribution $\mathcal{N}(\mu_{q|d}, K_{q|d})$ as follows:
\begin{subequations}
\label{eq:GP inference}
    \begin{align}
        \mu_{q| d} &= K_{q,d} (K_{d, d}+\nu^2 I)\inverse Y_{d},\\
        K_{q| d}=K_{q,q}| K_{d, d} &= K_{q,q}-K_{q,d} (K_{d, d}+\nu^2 I)\inverse K_{q,d}\transpose,
    \end{align}
\end{subequations}

where $K_{q,d}=k(X_q,X_d)$, $K_{q,q}=k(X_q,X_q)$. In order to safeguard the numerical stability of the covariance matrix during inference, we add a small jitter term $\nu^2 I$ (where $10^{-12}\leq\nu^2\leq10^{-9}$) to the diagonal of the kernel matrix. This perturbation ensures that the matrix remains well-conditioned and invertible in practical implementations %under finite-precision arithmetic . In the context of Gaussian Process emulators for deterministic computer models, the addition of the so-called ‘nugget’ term has been shown to mitigate ill-conditioning and improve numerical robustness
while having negligible impact on the predictive mean %when $\nu^2$ is kept sufficiently small
\cite{andrianakis2012effect}.

The value $\mu_{q\mid d}$ is typically used as a prediction of the value of $f(X_q)$ given the dataset $D$. GPs have two advantages over other regression techniques: first, the predicted covariance $K_{q\mid d}$ can serve as a measure of prediction confidence; second,  predictions are computed using pure linear algebra. However, a major limitation of GPs is given by the scaling of the computational complexity for inference with respect to dataset size, which is limited by the matrix inversion, which is greater than $O(n^2)$ \cite{tveit2003complexity}, independently of whether the computation is performed online or offline.

\subsection{Gaussian Processes with Gradient Information}
\label{sec:GPGintro}

The basic GP inference in \eqref{eq:GP inference} uses only point-wise information; however, since directional derivatives are linear operations, and Gaussian distributions remain Gaussian under linear transformations, the directional derivative of a GP is still a GP; moreover the kernel for this larger GP can be obtained from the directional derivatives of the original kernel \cite{McHutchon2013DifferentiatingGP,eriksson2018scaling}. We can therefore incorporate gradient information into a GP to improve accuracy. One point that needs attention is the fact that although $f(x)$ is a scalar process, its gradient is multi-dimensional; it is therefore necessary to consider correlations between the functions and each one of the partial derivatives.\footnote{At a first read, it might be helpful to the reader to consider the content of the section for the particular case of $d=1$.}

% However, we are dealing with an 8-dimensional input space $x \in \mathbb{R}^{8}$ in quadrotor dynamics, where 7 of these input dimensions have corresponding derivative information $\mathcal{J}$. Therefore, we extend the RBF and its partial derivative forms from the previous discussion to accommodate multivariate gradient information.

More formally, we use the notation
\begin{equation}
  \label{eq:partial def}
  \partial_n f= \frac{\partial f}{\partial [x]_n}, n\in\{1,\ldots,d\}
\end{equation}

to denote the partial derivative of $f$ with respect to the $n$-th element of $x$, $[x]_n$ to denote the $n$-th element of vector $x$, and $\partial f=\operatorname{stack}(\{\partial_n f\}_{n=1}^d)$ to denote the gradient of $f$.
\begin{remark}
The discussion in this section can be extended to the case where different directional derivatives are available at different points; the results would also apply to derivatives in arbitrary directions (not necessarily along the principal coordinate directions). The extension, however, would significantly complicate the notation and is omitted here for clarity.
\end{remark}

To derive the definition of the GP with gradient information, we need to first derive the covariance terms between the function and derivatives, as well as the covariance between derivatives. We start by redefining (for the sake of uniform notation) the base kernel as $k_{00}(x,x')=k(x,x')$.
Next, the kernel between the function and the derivative, and vice versa, are given by taking the partial derivatives of the kernel with respect to the corresponding argument:\begin{equation}
  \begin{aligned}
    k_{n0}(x,x') &= \frac{\partial k_{00}(x,x')}{\partial [x]_n}
      = -\frac{[x]_n - [x']_n}{l^2}k(x,x'),\\
    k_{0m}(x,x') &= \frac{\partial k_{00}(x,x')}{\partial [x']_{m}}
      = \frac{[x]_m - [x']_m}{l^2}k(x,x'),
  \end{aligned}
\end{equation}

where $n, m\in\{1,\ldots,d\}$. Finally, the kernel between derivatives is given by the second derivative of the kernel:
\begin{align}
  &k_{nm}(x,x') = \frac{\partial^2 k_{00}(x,x')}{\partial [x]_n\partial [x']_m} \\
      &= \begin{cases}
        \displaystyle \frac{l^2 - \bigl([x]_n - [x']_n\bigr)^2}{l^4}\;k(x,x'),
          & \text{if } n=m,\\
        \displaystyle -\tfrac{([x]_n - [x']_n)\;([x]_m - [x']_m)}{l^4}\;k(x,x'),
          & \text{if } n \neq m.
      \end{cases}
  \label{eq:kernel_gradient}
\end{align}

% where $K_{11}$ is the covariance matrix constructed from the derivative of the kernel function in all possible combinations within $X_g$:
% \begin{equation}
%     k_{11}(x_{g,i}, x_{g,j}) = \frac{\partial^2 k_{00}(x_{g,i} ,x_{g,j})}{\partial x_{g,i} \partial x_{g,j}}
% \end{equation}
% And $K_{10}$ and $K_{01}$ are the covariance matrices constructed from the derivative of the kernel function in all possible combinations between $X_g$ and $X_d$:
% \begin{equation}
%     \begin{aligned}
%         k_{10}(x_{d,i}, x_{g,j}) &= \frac{\partial k_{00}(x_{d,i}, x_{g,j})}{\partial x_{d,i}} \\
%         k_{01}(x_{d,i}, x_{g,j}) &= \frac{\partial k_{00}(x_{d,i}, x_{g,j})}{\partial x_{g,j}} \\
%     \end{aligned}
% \end{equation}

We are now ready to give the definition of the GP with gradient information.
In addition to the set of locations $X_d$ introduced in \Cref{sec:GPintro} for which function information is available, consider a dataset of $n_g$ locations $X_g=\operatorname{stack}(\{x_{g,i}\}_{i=1}^{n_g})$ for which the gradient information $\partial f$ is available. We define the full set of locations as $X_a=X_d\union X_g$.

We introduce the vector of random variables $f_a=\operatorname{stack}(f(X_d),\partial f(X_g))$, where the second term is organized as $\partial f(X_g)=\operatorname{stack}(\{\partial_nf(X_g)\}_{n=1}^{d})\in\mathbb{R}^{d\times n_g}$ (i.e., partial derivatives are organized by directions first, and then points).
\begin{remark}
  There is some flexibility in this formulation. First, $X_g$ can differ from $X_d$ (for the simulations in \Cref{sec:realtime_simulation}, we have $X_g=X_d$, but the different notation helps elucidate the relation between the function and gradient information). Second, the ordering of partial derivatives and points in $\partial f(X_g)$ can be arbitrary (we have selected the one above to ease the expression of the GP covariance).
\end{remark}

Under the assumption that the mean is zero, $f_a$ follows a Gaussian distribution:\begin{equation}\label{eq:fa distribution}
  f_a \sim \mathcal{N}(0, {\bar{K}}(X_{a}, X_{a})),
\end{equation}

where the covariance matrix is given by
\begin{multline}\label{eq:barK}
{\bar{K}}(X_{a}, X_{a}) =\\
  \begin{bmatrix}
    k_{00}(X_d,X_d) & k_{01}(X_d,X_g) & \cdots & k_{0d}(X_d,X_g)\\
    k_{10}(X_g,X_d) & k_{11}(X_g,X_g) & \cdots & k_{1d}(X_g,X_g)\\
    \vdots & \vdots & \ddots & \vdots\\
    k_{d0}(X_g,X_d) & k_{d1}(X_g,X_g) & \cdots & k_{dd}(X_g,X_g)
  \end{bmatrix}.\\
\end{multline}

\paragraph{Inference from training data}

Let $D_g=\{X_g, Y'_g\}$ be a dataset where $Y'_g$ is a realization (measurement) of $\partial f(X_g)$.
We define the combined dataset $D_a=\{X_a,Y_a\}$ where  $Y_a=\operatorname{stack}(Y_d, Y'_g)$. Given that the model follows the Gaussian distribution \eqref{eq:fa distribution}, regression in GPs with gradient information is analogous to regression in standard GPs, albeit with more information; specifically, given a set of query points $X_{q}$, the conditional distribution of $f(X_q)$ given $Y_a$ is a Gaussian distribution $\cN(\mu_{q\mid a}, K_{q\mid a})$ as follows:

\begin{subequations}
\label{eq:GP prediction gradient}
    \begin{align}
      \mu_{q\mid a} &=\bar{K}_{q,a}\bar{K}_{a,a}\inverse Y_{a}, \label{eq:GPGmean}\\
      K_{q\mid a}&=K_{q,q}-\bar{K}_{q,a} \bar{K}_{a,a}\inverse \bar{K}_{q,a}\transpose, \label{eq:GPGvariance}
    \end{align}
\end{subequations}

where
\begin{equation}
\label{eq:gradient_covariance_noise}
  \begin{aligned}
    \bar{K}_{q,a} &= \bar{K}(X_q,X_a)
    \\&=\begin{bmatrix}
        k_{00}(X_q,X_d) & k_{01}(X_q,X_g) & \dots & k_{0d}(X_q,X_g)
    \end{bmatrix},\\
    \bar{K}_{a,a} &=\\
        &\begin{bmatrix}
            A_{00} & k_{01}(X_d,X_g) & \cdots & k_{0d}(X_d,X_g)\\
            k_{10}(X_g,X_d) & B_{11} & \cdots & k_{1d}(X_g,X_g)\\
            \vdots & \vdots & \ddots & \vdots\\
            k_{d0}(X_g,X_d) & k_{d1}(X_g,X_g) & \cdots & B_{dd}
        \end{bmatrix},\\
        A_{00} &= k_{00}(X_d,X_d) + \nu^2 I,\\
        B_{ii} &= k_{ii}(X_g,X_g) + \lambda_i^2 I,\;\; i\in\{1\ldots d\}.
  \end{aligned}
\end{equation}

Here, the parameters $\lambda_i^2$ capture the noise variance of the $i$-th dimension's of $Y'_g$ \cite{Rasmussen2006Gaussian}; this information models the fact that, in practice, even simulation-based training data may exhibit residual variability, modeling error or approximations (e.g., the use of finite differences to compute derivatives), or actuator/solver noise; tuning of this parameter allows us to avoid overly confident predictions from the GP model.
\begin{remark}
  It is possible to obtain similar expressions for the distribution of $\partial f(X_q)$ given $Y_a$, but these are not used in the application considered in this paper.
\end{remark}
\begin{remark}\label{remark:no gradient}
  If the gradient information for a certain input dimension is not available, we just need to omit the corresponding rows/columns in $\bar{K}_{a, a}$ and $Y'_g$ for that dimension.
\end{remark}
\subsection{Simple Quadrotor Aerodynamic Model}
\label{sec:SimpleModel}

In this study, we employ the NASA SUI Endurance quadrotor UAV for data collection and simulation target. This platform features four rotors arranged in a square configuration on the same horizontal plane. Accordingly, we describe its simplified dynamics model $\mathcal{M}$ of forces and torques in the body-fixed coordinate frame $\mathcal{B}$ as follows \cite{mellinger2011minimum}:
\begin{equation}
    \begin{aligned}
        \prescript{\mathcal{B}}{}{F}_{\mathcal{M}} = \begin{bmatrix}
            0&0&P_F \sum^{4}_{n=1}({r_{n}}^2)
        \end{bmatrix}\transpose,\\
        \prescript{\mathcal{B}}{}{\tau}_{\mathcal{M}}= \begin{bmatrix}
            l_xP_F(-{r_1}^2+{r_2}^2+{r_3}^2-{r_4}^2) \\
            l_yP_F(-{r_1}^2-{r_2}^2+{r_3}^2+{r_4}^2) \\
            T_z(-{r_1}^2+{r_2}^2+{r_3}^2-{r_4}^2)
        \end{bmatrix},
    \end{aligned}
\end{equation}

where $l_x$ and $l_y$ denote the distances from the rotor centers to the respective coordinate axes, measured in meters, and $T_z$ is a constant with units of $\unitfrac{N\cdot m}{rpm^2}$, obtained from previous experimental data for this UAV platform.

The equations above represent the simplest dynamics model of the quadrotor. In these equations, $r_n$ denotes the rotational speed of the corresponding rotor, measured in revolutions per minute (\unit{rpm}). $P_F$ is the thrust coefficient of the rotor, with units of $\unitfrac{N}{rpm^2}$, determined from the known condition that the SUI Endurance achieves stable hover at around $\unit[3500]{rpm}$.
Next, we define the sequence of rotations that converts forces $\Bframe{F}_{\mathcal{M}}$ and torques $\prescript{\mathcal{B}}{}{\tau}_{\mathcal{M}}$ from the body-fixed frame $\mathcal{B}$ to the world frame $\mathcal{W}$. The process follows the order of yaw $\psi$, pitch $\theta$, and roll $\phi$, and can be expressed as follows:
\begin{align}
    % R_z(\psi) &=
    % \begin{bmatrix}
    % \cos\psi & -\sin\psi & 0 \\
    % \sin\psi & \cos\psi  & 0 \\
    % 0        & 0         & 1
    % \end{bmatrix} \\
    % R_y(\theta) &=
    % \begin{bmatrix}
    % \cos\theta & 0 & \sin\theta \\
    % 0          & 1 & 0 \\
    % -\sin\theta& 0 & \cos\theta
    % \end{bmatrix}\\
    % R_x(\phi) &=
    % \begin{bmatrix}
    % 1 & 0        & 0 \\
    % 0 & \cos\phi & -\sin\phi \\
    % 0 & \sin\phi & \cos\phi
    % \end{bmatrix}\\
    \prescript{\mathcal{W}}{}{R}_{\mathcal{B}} &= R_z(\psi) R_y(\theta) R_x(\phi),\\
    % &=\begin{bmatrix}
    % \cos\psi\cos\theta &
    % \cos\psi\sin\theta\sin\phi - \sin\psi\cos\phi &
    % \cos\psi\sin\theta\cos\phi + \sin\psi\sin\phi \\
    % \sin\psi\cos\theta &
    % \sin\psi\sin\theta\sin\phi + \cos\psi\cos\phi &
    % \sin\psi\sin\theta\cos\phi - \cos\psi\sin\phi \\
    % -\sin\theta &
    % \cos\theta\sin\phi &
    % \cos\theta\cos\phi
    % \end{bmatrix}\\
    \prescript{\mathcal{W}}{}{F}_{\mathcal{M}} &= \prescript{\mathcal{W}}{}{R}_{\mathcal{B}}\prescript{\mathcal{B}}{}{F}_{\mathcal{M}},\\
    \prescript{\mathcal{W}}{}{\tau}_{\mathcal{M}} &= \prescript{\mathcal{W}}{}{R}_{\mathcal{B}}\prescript{\mathcal{B}}{}{\tau}_{\mathcal{M}},\\
    y_\mathcal{M} &= \operatorname{stack}(\prescript{\mathcal{W}}{}{F}_{\mathcal{M}}, \prescript{\mathcal{W}}{}{\tau}_{\mathcal{M}}),
\end{align}

where $R_x$, $R_y$, and $R_z$ are rotation matrix calculated from the corresponding angles. $\prescript{\mathcal{W}}{}{R}_{\mathcal{B}}$ denotes the rotation matrix from $\mathcal{B}$ to $\mathcal{W}$. Since these predictions used the rpms of the propellers $\mathbf{r}=\operatorname{stack}(r_1,r_2,r_3,r_4 )$ and the rotation angles $\Theta =\operatorname{stack}(\psi,\theta,\phi)$, we define the quadrotor state used by $\mathcal{M}$ in world frame $\mathcal{W}$ as $\prescript{\mathcal{W}}{}{x}_\mathcal{M}=\operatorname{stack}(r_1, r_2, r_3, r_4, \psi,\theta,\phi)$.

As our dataset also contains gradient information that describes the change rate of $\prescript{\mathcal{W}}{}{F}_{\mathcal{M}}$ and $\prescript{\mathcal{W}}{}{\tau}_{\mathcal{M}}$ with respect to $\prescript{\mathcal{W}}{}{x}_\mathcal{M}$. We will also need these from our simplified model to rule out the part predicted by it. We denote the Jacobian of $(\prescript{\mathcal{W}}{}{F}_{\mathcal{M}}, \prescript{\mathcal{W}}{}{\tau}_{\mathcal{M}})$ with respect to the 7-dimensional input $\prescript{\mathcal{W}}{}{x}_\mathcal{M}$ by $\prescript{\mathcal{W}}{}{J}_\mathcal{M}\in\mathbb{R}^6\times\mathbb{R}^7$. Its $j$-th column is the gradient of the outputs with respect to the $j$-th input element.
% The gradients calculated by the model $\mathcal{M}$ in the world frame $\mathcal{W}$ will be denoted as $\prescript{\mathcal{W}}{}{J}_\mathcal{M}=\operatorname{stack}(\{\partial_iy_\mathcal{M}\transpose\}_{i=1}^7)$. In general, the column of a Jacobian is the gradient of the outputs with respect to the corresponding input.
Since the derivation of the gradient expressions is straightforward, we omit the explicit formulation of these gradient values here.
% Given that the input of the simplified model is $\mathbb{R}^{7}$, and the output is $\mathbb{R}^{6}$, the resulting Jacobian matrix has a dimension of $\prescript{\mathcal{W}}{}{J}_\mathcal{M}\in \mathbb{R}^{6}\times\mathbb{R}^{7}$.

Notice that this simple model does not take into account airspeed $v$ nor provide predictions of the quadrotor noise level $L_\mathcal{M}$. These two factors will be discussed in detail in \Cref{sec:data_collection}.

For clarity in the subsequent discussion, we denote the simplified quadrotor dynamics model for the calculation of forces, torques, and the corresponding Jacobian matrix for a given  quadrotor state $x_\mathcal{M}$ as:
\begin{equation}
\label{eq:simple_model_prediction}
    \mathcal{M}(\prescript{\mathcal{W}}{}{x}_\mathcal{M})= \operatorname{stack}(\prescript{\mathcal{W}}{}{F}_\mathcal{M}, \prescript{\mathcal{W}}{}{\tau}_\mathcal{M}, \prescript{\mathcal{W}}{}{J}_\mathcal{M}).
\end{equation}

\section{Partitioned GP with Gradient Information}
\label{sec:partition}
Our objective is to develop an accurate, mid-fidelity quadrotor dynamics simulation environment that runs in real time on commodity hardware.

The products involved in the GP prediction \eqref{eq:GP prediction gradient} can be categorized into two types:
\begin{enumerate*}
    \item \emph{precomputable} (such as $\bar{K}_{a,a}\inverse Y_a$), which depend solely on the dataset $D_a$, and can therefore be computed offline, and
    \item \emph{query-dependent} (such as $K_{q,q}$, $\bar{K}_{q,a}\bar{K}_{a,a}\inverse \bar{K}_{q,a}\transpose$), which involve the query point $x_q$ and must be computed in real time.
\end{enumerate*}

Since precomputable products do not affect runtime during prediction, our focus is on improving the computational efficiency of query-dependent products.

We introduce a \emph{partitioned GP architecture}, which allows us to consider only the computation of correlations between the query point $x_q$ and a \emph{near set} $X_{q, n} \subset X_d$, neglecting the set of points $X_{q,f}=X_d \setminus X_{q,n}$. With respect to previous work, we introduce two modifications. First, we use an edge-condition analysis for each (\Cref{sec:Binning}) that transforms the determination of $X_{q,n}$ into a fast look-up operation. Second, we apply \emph{Schur complement-based corrections} for the covariance and the mean (\Cref{sec:SchurPredict}), which reintegrate the influence of $X_{f}$ into the local prediction.

\subsection{Partitioning (Selection of Near and Far Sets)}
\label{sec:Binning}

A na\"ive approach to the computation of near and far sets would require computing the correlation matrix $\bar{K}_{q, a}$, and then discard points in $a$ that correlation lower than a user-defined threshold; this, however, requires computing, online, the correlation between each query and \emph{all} the points in the dataset, an operation that is itself computationally expensive.

A more common approach is to partition the input space into bins (e.g., via clustering) so that each bin contains approximately the same number of data points. For each query $x_{q}$, then, only the points $X_{n}$ in the same bin are considered, completely discarding the remaining points $X_f$.

This basic partition creates two problems:\begin{enumerate*}
    \item when $x_{q}$ is close to the boundary of the bin, the prediction is likely to discard points that are in another bin, but still have a significant correlation with the query;
    \item the loss of information where, other than the local sub dataset, other blocks are discarded.
\end{enumerate*}

% Thus, we want to segment the dataset into overlapping blocks. The blocks of data should be covering each others so the boundary condition in one block should be the center condition in another block. When receiving a $x_{query}$, we can always having a $X_{near}$ where $x_{query}$ is closer to the center than to the boundary.

\begin{figure}[tp]
    \centering
    \includegraphics[width=\linewidth]{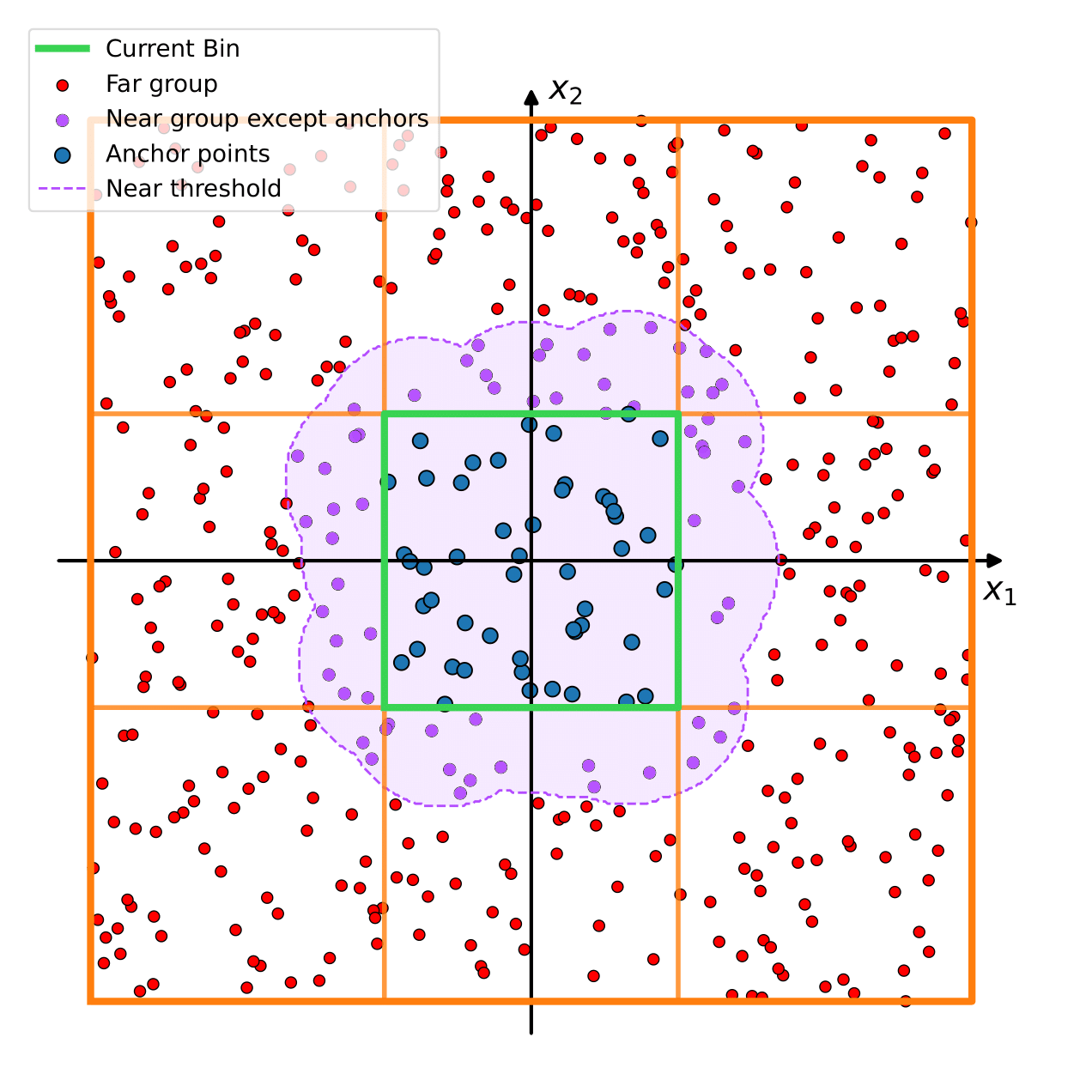}
    \caption{Diagram for partitioned Gaussian Process}
    \label{fig:binning}
\end{figure}

Instead of limiting data selection to mutually exclusive bins, we expand the selection to overlapping areas based on correlation. For each bin $B=\{ b_1, b_2, b_3, \dots\}$, we identify anchor points $X_{b_i}=\operatorname{stack}( x_{b_i,1}, x_{b_i,2}, x_{b_i,3},\dots )$ as data points within the range of the bin (see \Cref{fig:binning} for an example). For a bin $b_i$, we classify the entire dataset $X_d$ into two groups based on the correlation given by the kernel function: if the correlation is above a specified threshold $\epsilon$, we consider the point to be in the near set $X_{n_i}$, otherwise it is assigned to the far set $X_{f_i}$. This grouping method is inspired by the well-established clustering method DBSCAN \cite{ester1996density}.
This conservative classification ensures that for any state within the block, its kernel correlation with $X_{n_i}$ generally remains above the threshold $\epsilon$, while its correlation with $X_{f_i}$ definitely remains below $\epsilon$. This partition method is the foundation for our work in \Cref{sec:SchurPredict}.

\subsection{Schur Complement approximation of prediction}
\label{sec:SchurPredict}

Since we are grouping the dataset, the covariance matrix $\bar{K}_{a,a}$ can be partitioned into blocks similar to \eqref{eq:SchurInv}. Following this Schur complement idea, the full prediction of the covariance in \eqref{eq:GPGvariance} can be written, with some reordering of the points and gradient entry to group as $\bar{X}_{n_i}$ and $\bar{X}_{f_i}$ for bin $b_i$, as:\begin{multline}
    K_{q|a} = K_{qq}-
    \begin{bmatrix}
        \bar{K}_{q,n_i} \\ \bar{K}_{q,f_i}
    \end{bmatrix}\transpose
     \begin{bmatrix}
         \bar{K}_{n_i,n_i} & \bar{K}_{n_i,f_i} \\ \bar{K}_{f_i,n_i} & \bar{K}_{f_i,f_i}
     \end{bmatrix}\inverse
     \begin{bmatrix}
         \bar{K}_{q,n_i} \\ \bar{K}_{q,f_i}
     \end{bmatrix} \\
    =K_{qq}-
    \begin{bmatrix}
        \bar{K}_{q,n_i} \\ \bar{K}_{q,f_i}
    \end{bmatrix}\transpose
    \begin{bmatrix}
        (\bar{K}_{a,a}|\bar{K}_{f_i,f_i})\inverse & * \\ * & *
    \end{bmatrix}
    \begin{bmatrix}
        \bar{K}_{q,n_i} \\ \bar{K}_{q,f_i}
    \end{bmatrix},
    \label{eq:SchurCov_rearrange}
\end{multline}

where we used Schur complements for the matrix inversion of $\bar{K}_{a, a}$. The entries marked as $*$ are given by the usual Schur complement formula for matrix inversion. In the above equation, we have $\bar{K}_{n_i,f_i}=\bar{K}(\bar{X}_{n_i},\bar{X}_{f_i})$, $\bar{K}_{n_i,n_i}=\bar{K}(\bar{X}_{n_i},\bar{X}_{n_i})$, and $\bar{K}_{f_i,f_i}=\bar{K}(\bar{X}_{f_i},\bar{X}_{f_i})$.

A similar rearrangement in \eqref{eq:SchurCov_rearrange} can also be applied to \eqref{eq:GPGmean}:
\begin{multline}
    \mu_{q|a,b_i} =
    \begin{bmatrix}
        \bar{K}_{q,n_i} \\ \bar{K}_{q,f_i}
    \end{bmatrix}\transpose
     \begin{bmatrix}
         \bar{K}_{n_i,n_i} & \bar{K}_{n_i,f_i} \\ \bar{K}_{f_i,n_i} & \bar{K}_{f_i,f_i}
     \end{bmatrix}\inverse
     \begin{bmatrix}
         \bar{Y}_{n_i} \\ \bar{Y}_{f_i}
     \end{bmatrix} \\
    =
    \begin{bmatrix}
        \bar{K}_{q,n_i} \\ \bar{K}_{q,f_i}
    \end{bmatrix}\transpose
    \begin{bmatrix}
        (\bar{K}_{a,a}|\bar{K}_{f_i,f_i})\inverse & * \\ * & *
    \end{bmatrix}
    \begin{bmatrix}
        \bar{Y}_{n_i} \\ \bar{Y}_{f_i}
    \end{bmatrix}.
    \label{eq:SchurMean_rearrange}
\end{multline}

But due to the threshold $\epsilon$ we set in \Cref{sec:Binning}, we can approximate $\bar{K}_{q,f_i}$ to $0$. This makes $*$ entries in \eqref{eq:SchurCov_rearrange} and \eqref{eq:SchurMean_rearrange} irrelevant, so we can write these equations as:

% \begin{equation}
%     K_{q|a,b_i}= K_{qq}-\bar{K}_{q,n_i}(\bar{K}_{a,a}|\bar{K}_{f_i,f_i})\inverse \bar{K}_{q,n_i}\transpose \label{eq:prediction covariance approx}
% \end{equation}
\begin{subequations}
\label{eq:GPGS_prediction}
    \begin{align}
      \mu_{q|a,b_i}&=\bar{K}_{q,n_i}(\bar{K}_{a,a}|\bar{K}_{f_i,f_i})\inverse \bar{Y}_{n_i}, \label{eq:GPGSmean}\\
      K_{q|a,b_i}&= K_{qq}-\bar{K}_{q,n_i}(\bar{K}_{a,a}|\bar{K}_{f_i,f_i})\inverse \bar{K}_{q,n_i}\transpose. \label{eq:GPGSvariance}
    \end{align}
\end{subequations}

The Schur complement is given by:
\begin{equation}
    \bar{K}_{a,a} | \bar{K}_{f_i,f_i} = \bar{K}_{n_i,n_i} - \bar{K}_{n_i,f_i} \bar{K}_{f_i,f_i}\inverse \bar{K}_{n_i,f_i}\transpose.
    \label{eq:prediction covariance schur}
\end{equation}

Using the Schur complement, we adjust $\bar{K}_{n_i, n_i}$ based on the information from $\bar{X}_{f_i}$. This also helps us save computational cost by reducing the covariance calculation in real-time simulation from $\bar{K}_{q, a}$ to $\bar{K}_{q, n_i}$.

A direct computation of \eqref{eq:prediction covariance schur} requires the inversion of $\bar{K}_{f_i,f_i}$, which is almost as large as the original $\bar{K}_{a, a}$. However,
\begin{itemize}
\item If we fix the $near$, $far$ selection (by overapproximating the neighbors of $q$ with fixed overlapping groups), then all the $\bar{K}$ in \eqref{eq:prediction covariance schur} can be computed offline.
\item We can compute $\bar{K}_{f_i, f_i}\inverse$ using an iterative solver (PyTorch) (for as long as necessary) and store the result.
\end{itemize}

\section{Dataset Reduction through Gradient Information}

To further lower the \emph{query-dependent} calculation ($K_{q,q}$, $\bar{K}_{q,a}\bar{K}_{a,a}\inverse \bar{K}_{q,a}\transpose$), we also want to shrink the dataset size $n_a$ required for an accurate prediction. We utilize \emph{gradient information} (\Cref{sec:GPGintro}) to reduce the number of base points $\bar{X}_a$ required to achieve the desired prediction accuracy.

% During data collection, we use numerical differentiation to merge 8 data points into a single point with gradient information across 7 dimensions.

Although the correlation with the gradient observation still requires calculation, it should be noted that in \eqref{eq:kernel_gradient}, both the first- and second-order derivatives of the selected GP kernel function can be obtained by multiplying the original kernel function by the corresponding differences between input dimensions. These different terms are already computed when evaluating the original kernel function itself. This implies that, compared to computing a new kernel value between an entirely different pair of points, the additional cost of evaluating the first- or second-order derivatives of the kernel is relatively small.

% This reduction in computational load provides a speed advantage when incorporating gradient information into the GP. During real-time prediction, computing the correlations between a query point and a smaller dataset enriched with derivative information is significantly faster than evaluating correlations with a larger dataset of raw samples.

\section{Aerodynamic Surrogate}
\label{sec:aero_modeling}

In this section, we show the pipeline of our work: \begin{itemize}
    \item Using CHARM and PSU-WOPWOP to generate the simulation data.
    \item Modification of the dataset based on the simplified dynamics model and normalization of the residuals.
    \item Partitioning and precomputation of Schur-based approximation.
    \item Real-time simulation process.
\end{itemize}

The following subsections cover each step of the pipeline, detailing how it is used to model the forces $F$ and torques $\tau$ applied to the center of mass of the quadrotor, along with noise levels $L$ sampled at three points, located in the body frame, around the quadrotor.

\subsection{Aerodynamic and Acoustic Simulation}
%The Comprehensive Hierarchical Aeromechanics Rotorcraft Model (CHARM) developed by Continuum Dynamics, Inc. \cite{CHARM}

%It builds on the fundamental laws of physics: conservation of mass, momentum, and energy.

%Instead of doing an experiment with a wind tunnel, we model the object and flow field on a computer and solve the equations numerically.

\begin{figure}[h!]
    \centering
    \includegraphics[width=1.0\linewidth]{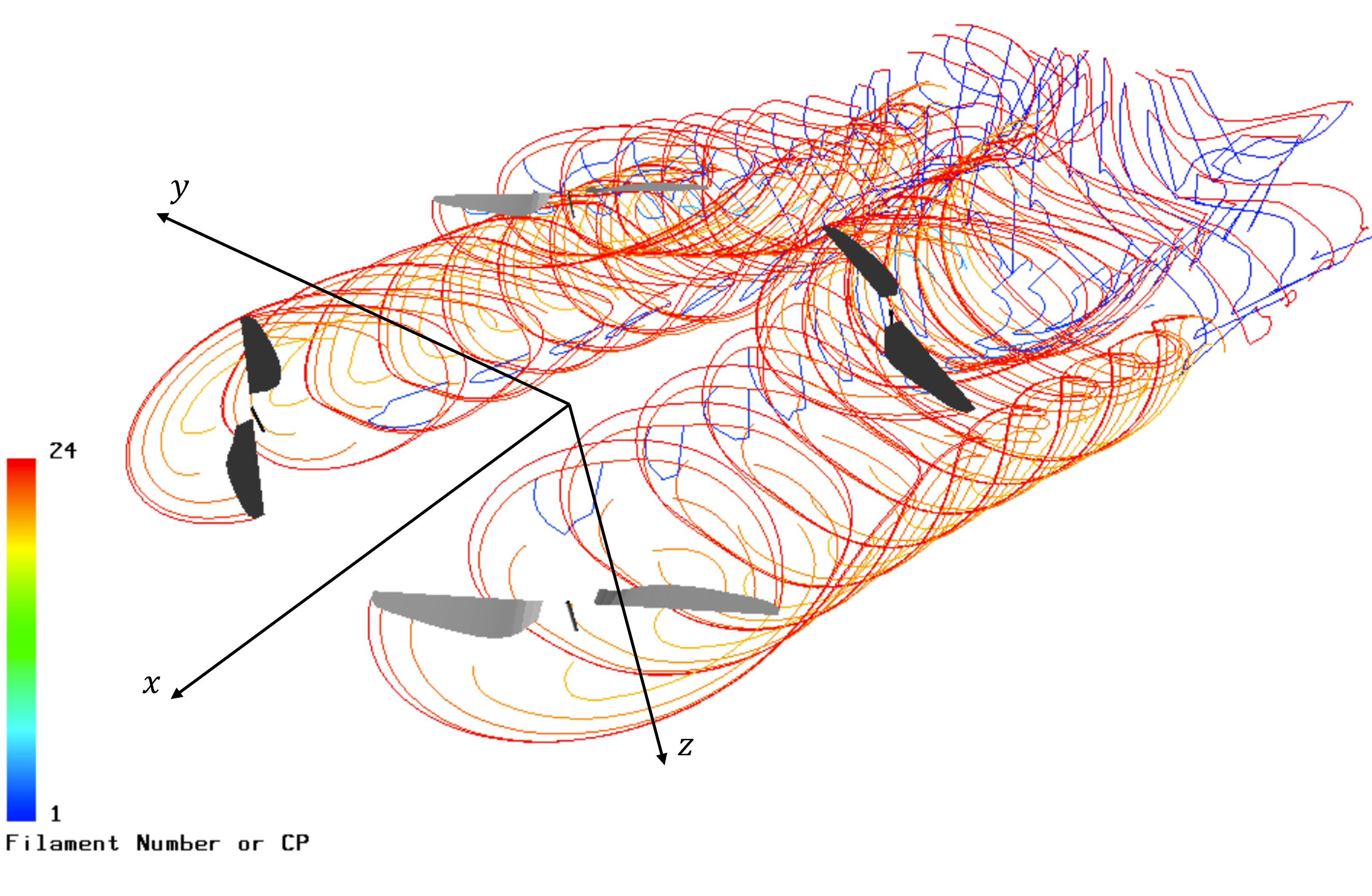}
    \caption{CHARM Quadrotor Solution Wake Visualization}
    \label{fig:charm_wake}
\end{figure}

CHARM's solution method is rooted in potential flow analysis, which greatly simplifies the governing fluid equations by ignoring viscosity. Potential flow tools utilize the linearity of the potential equations to create solutions that are superpositions of elementary flows. The CHARM tool couples a full-span, free-vortex wake model with a vortex lattice lifting line model on the blades. A sample output visualizing the wake calculation is visible in \Cref{fig:charm_wake}.

Through the lifting line model, viscous effects are added back in. The lifting line relies on airfoil tables, which provide the aerodynamic information for each slice of the propeller blade (i.e. lift, drag, and moment). Therefore, the method used to develop the airfoil table influences aerodynamic predictions. For this work, the airfoil table was developed at NASA using NASA's Fun3d, a finite volume Reynolds Averaged Navier Stokes (RANS) solver, and the Spalart Allmaras (SA) turbulence model \cite{russellsekula}.
CHARM predictions have been validated against experimental measurements for several multirotor configurations \cite{shirazi_multirotor_CHARM, botre_PSU_CHARM}. Russel et al. \cite{russell2016wind} measured performance values for the SUI quadrotor discussed in this work and Zawodny et al. \cite{zawodny2018acoustic} published noise measurements for the drone in several configurations in the NASA Langley Low Speed Acoustic Wind Tunnel.

\begin{figure}[t]
    \centering
    \includegraphics[width=1.0\linewidth]{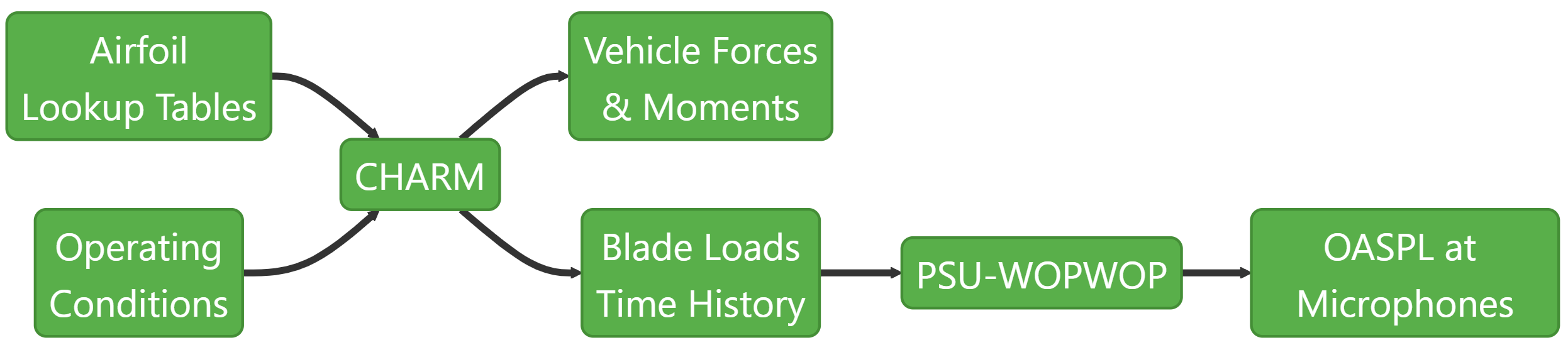}
    \caption{Workflow for a CHARM Aerodynamic Simulation}
    \label{fig:charm_diag}
\end{figure}

CHARM is readily coupled to an acoustic propagation tool, PSU-WOPWOP \cite{botre_PSU_CHARM}.
CHARM  writes chordwise-integrated loads as well as a basic geometry file to PSU-WOPWOP \cite{WOPWOP} input format. The workflow is summarized in \Cref{fig:charm_diag}.

CHARM does not model turbulence, so only tonal noise can be predicted. The use of CHARM with PSU-WOPWOP to predict tonal noise for propellers has been demonstrated \cite{mukherjee,truong,marques}. A new method for utilizing PSU-WOPWOP to isolate tones associated with a multirotor vehicle like the SUI is presented in \cite{SUI_scitech26}, which shows that the methods used in this paper compare well to experimental data.

\subsection{Data Collection}
\label{sec:data_collection}

%Training data was generated using the Comprehensive Hierarchical Aeromechanics Rotorcraft Model (CHARM) developed by Continuum Dynamics, Inc. \cite{CHARM}. CHARM employs the vortex lattice method which involves dividing bodies such as propeller blades into panels, seeding and tracking discrete vortex elements to model wakes, and applying forces to the wake elements according to airfoil lookup tables. Viscous effects such as boundary layers are not modeled. This modeling approach captures the wake physics of multirotor vehicles and produces performance and acoustic results for low computational cost.

%\textcolor{red}{If it makes sense for us to supply a sample set of cases -- am sending the abstract - we can do this.  We can discuss the NASA performance test cases and show how well the CHARM does in getting the performance. Then we can share a figure like what we will have in the Scitech paper. }

The dataset was generated using Latin Hypercube (LHC) sampling \cite{LHC} of airspeed, individual propeller RPM, and vehicle yaw, pitch, and roll scaled according to the bounds in \Cref{tab:LHC_values_booktabs}.
\begin{table}[h!]
\centering % Use \centering instead of the center environment
\caption{Dataset LHC Sampling Bounds}
    \label{tab:LHC_values_booktabs}\begin{tabular}{l c c} % 'l' for left-aligning the first column
    \toprule % Top rule
    \textbf{Variable} & \textbf{Lower Limit} & \textbf{Upper Limit} \\
    \midrule % Middle rule
    Airspeed (m/s) & 0 & 20 \\
    Propeller Rotation Speed (RPM) & 0 & 7000 \\
    Yaw (deg) & -90 & 90 \\
    Pitch (deg) & -60 & 60 \\
    Roll (deg) & -90 & 90 \\
    \bottomrule % Bottom rule
\end{tabular}
\end{table}
The simulation environment is setup as follows: the quadrotor is always moving in the $+x$ direction of the world frame. The $+z$ direction of the world frame points downward, and the $+y$ direction follows the right-hand rule (see the reference frame visualization in \Cref{fig:charm_wake}). %We adjusted the Euler angles to get different sample points.

A CHARM case was run for each sample point to extract forces and moments. Reduced-order schemes were used, allowing each case to complete in less than one minute using a single CPU core.
Acoustic processing was performed to extract only the amplitudes of the first ten multiples of the Blade Passage Frequency ($\Omega\times n_{blades}$) of each propeller, which were then integrated to obtain an Overall Sound Power Level. This value was recorded at three microphones spaced equally along on an arc with radius $20\times R_{propeller}$ starting at the side, in the propeller plane, and ending directly below the vehicle. These values make up the noise output vector $L$ used in our GP model.

The derivatives at each point were calculated using finite difference by running a separate CHARM case with the propeller RPM of interest perturbed by 10\% or the roll, pitch, and yaw by one degree.

For a case $i$, we record the case state $x_i\in\mathbb{R}^8$, the aerodynamics part of the output value $y_{aero,i}\in\mathbb{R}^6$, the corresponding gradient information $J_{aero,i}\in\mathbb{R}^6\times\mathbb{R}^7$, the noise part of the output value $y_{noise,i}\in\mathbb{R}^3$, and the corresponding gradient information $J_{noise,i}\in\mathbb{R}^3\times\mathbb{R}^7$:
\begin{equation}
    \begin{aligned}
        x_i&=\stack(r_{i,1}, r_{i,2},r_{i,3},r_{i,4},\psi_{i},\theta_{i},\phi_{i},v_{i}), \\
        y_{\text{aero},i}&=\stack(F_{x,i},F_{y,i},F_{z,i},\tau_{x,i},\tau_{y,i},\tau_{z,i}),\\
        y_{\text{noise},i}&=\stack(L_{1,i},L_{2,i},L_{3,i}), \\
        J_{aero,i}&=
        \begin{bmatrix}
            \partial_1 F_{x,i} & \dots & \partial_7 F_{x,i}\\
            \vdots & \ddots & \vdots \\
            \partial_1 \tau_{z,i} & \dots & \partial_7 \tau_{z,i}
        \end{bmatrix},\\
        J_{noise,i}&=
        \begin{bmatrix}
            \partial_1 L_{1,i} & \dots & \partial_7 L_{1,i}\\
            \vdots & \ddots & \vdots \\
            \partial_1 L_{3,i} & \dots & \partial_7 L_{3,i}
        \end{bmatrix}.
    \end{aligned}
\end{equation}

Note that the gradient information is available only with respect to the four rotor speeds and three Euler angles (7 dimensions); we do not compute derivatives with respect to the airspeed (see \Cref{remark:no gradient}).

% Our dataset was primarily obtained through computational fluid dynamics (CFD) simulations. The input (state space) is defined over seven dimensions: the rotational speeds of the four rotors $\mathbf{r}$ and the three orientation angles of the UAV relative to the world coordinate frame $\Theta$. The output is defined over six dimensions: the forces acting on the UAV resolved along the three world axes $F_W$, and the torques resolved along the same axes $\tau_W$.

% Since gradient information can improve predictive accuracy beyond the sampled data domain, we also extracted gradient information of the dynamics $\mathcal{J}$ from the simulations using numerical differentiation. Specifically, by slightly perturbing a single input dimension and comparing the resulting changes across all outputs relative to the baseline values, we approximate the gradient as the ratio between input perturbation and the corresponding output variation.

\subsection{Data Processing}
\label{sec:data_processing}

To improve the prediction accuracy of the GP, we want to use the simple quadrotor dynamics model $\mathcal{M}$ (\Cref{sec:SimpleModel}), leaving the GP model to predict a correction term. % based on $\mathcal{M}$.
This correction term for each data point is calculated as $y_{\mathcal{M},i}=\stack(\prescript{\mathcal{W}}{}{F}_{\mathcal{M},i}, \prescript{\mathcal{W}}{}{\tau}_{\mathcal{M},i})$. The gradient information $J_{\mathcal{M},i}\in\mathbb{R}^6\times\mathbb{R}^7$ by computing the derivative of \eqref{eq:simple_model_prediction} (without considering terms related to either the acoustic noise or airspeed, which are not part of the model $\cM$).

We define the difference between the model calculation and the aerodynamic output and gradient as the force and torque correction terms that will modeled by the GP:
\begin{align}
    y_{c,i} &= y_{aero,i} - y_{\mathcal{M},i},\\
    J_{c,i} &= J_{aero,i} - J_{\mathcal{M},i}.
\end{align}

The correction terms for force and torque, combined with the original noise data $y_{noise, i}$ and its gradient $J_{noise, i}$, form the raw dataset for the Gaussian process:
\begin{equation}
    \begin{aligned}
        y_{i}&=\stack(y_{c,i},y_{noise,i})\in\mathbb{R}^9,\\
        J_{i} &=\stack(J_{c,i}, J_{noise,i})\in \mathbb{R}^9 \times \mathbb{R}^7, \\
        Y_d&=\stack(\{y_i\}_{i=1}^{n_a})\in\mathbb{R}^{n_a}\times\mathbb{R}^9.\\
        % \partial_j Y_d&=\stack(\{ [J_i]_j\transpose\}_{i=1}^{n_a}) \in \mathbb{R}^{n_a}\times\mathbb{R}^9,
    \end{aligned}
\end{equation}

To further improve prediction accuracy, we apply \emph{min-max normalization} to each dimension of the quadrotor state $x$ to eliminate magnitude differences between dimensions \eqref{eq:normalize_x}. Simultaneously, we perform \emph{z-score standardization} on each output dimension to align with the Gaussian Process model’s expectation of zero-mean unit-variance data \eqref{eq:normalize_y}. In addition, the same normalization parameters are applied to the gradient information, ensuring consistency between standardized states and outputs \eqref{eq:normalize_g}. The resulting dataset, after these processing steps, serves as the input to our Gaussian Process model:

\begin{subequations}
    \begin{align}
        X_d &= \stack(\{x_i\}_{i=1}^{n_a}),\\
        \tilde{X}_d &= 2\frac{X_d-\min(X_d)}{\max(X_d)-\min(X_d)}-1,\label{eq:normalize_x}\\
        \tilde{Y}_{d} &= \frac{Y_d - \mu(Y_d)}{\sqrt{Var(Y_d)}},\label{eq:normalize_y}\\
        \partial_j\tilde{Y}_d &= stack\left( \biggl\{ \frac{[J_i]_j[\frac{\max(X_d)-\min(X_d)}{2}]_j}{[\sqrt{Var(Y_d)}]_j}\transpose \biggl\} _{i=1}^{n_a}\right),\label{eq:normalize_g}\\
        \bar{Y}_a&=\stack(\tilde{Y}_d,\{\partial_j\tilde{Y}_d\}^7_{j=1}\}\in \mathbb{R}^{8n_a}\times\mathbb{R}^9,\\
        \bar{X}_a &= \stack(\{ \tilde{X}_d\}_{j=1}^8)\in\mathbb{R}^{8n_a}\times\mathbb{R}^{8},
    \end{align}
\end{subequations}

where, $[J_i]_j$ denotes the $j$-th column of the Jacobian matrix $J_i$; note that, as a result $\partial_j\tilde{Y}_d\in\mathbb{R}^{n_a}\times\mathbb{R}^9$ is defined as the matrix whose $i$-th row is the $j$-th column of regulated $J_i$ (i.e., the regulated gradient of all nine outputs with respect to $[x]_j$).

We compute the coefficients $\lambda_i^2$ for the regularization terms in \eqref{eq:gradient_covariance_noise} as
%After normalizing the gradient information, we will calculate the $\lambda$ for each input dimension as:
\begin{equation}
  \lambda_i^2 = s\times Var(\partial_i\tilde{Y}_{d}),
\end{equation}
where $s$ is a tuning parameter quantifying the errors in the simulation data; in practice, we use $s=0.3$.

%This standard deviation $\tilde{\lambda}_d=\stack(\lambda_1, \dots, \lambda_7)$ is collected to be used in . The gradient information in our dataset was generated numerically rather than analytically, as shown in \Cref{sec:data_collection}, so it is not entirely accurate. We treat this inaccuracy as noise in gradient information. By adjusting the percentage of the deviation we applied in \eqref{eq:gradient_covariance_noise}, we can adjust the model's confidence in the gradient information. In practice, we apply $30\%$ of the $\lambda_i$.

\subsection{Partitioning and precomputation of the Schur-based approximation}
\label{sec:precomputation}

Before running the real-time simulation, we need to perform some preparatory computations.

To achieve %an accurate enough physical simulation of the real world, we are targeting
an update frequency of at least \unit[$30$]{Hz} %for the simulation system. For an ordinary home desktop PC to perform the calculation of mean and variance at \unit[$30$]{Hz} in Python,
on commodity hardware (as the PC used for our simulations below), we found that the size of $K_{a,a}$ should be of the magnitude of $\mathbb{R}^{10^4}\times\mathbb{R}^{10^4}$ for a standard GP; after considering the gradient information dimensions, this corresponds to partitioning the state space into blocks $X_{n_i}$ of at approximately $10^3$ data points. %, to save space for the 7-dimensional gradient information.

For a given block $b_i$, we divide $X_a$ into near group $X_{n_i}$ and far group $X_{f_i}$ based on \Cref{sec:Binning}; we then we follow the calculation process in \Cref{sec:SchurPredict}, calculating and storing $(\bar{K}_{a,a}|\bar{K}_{f_i,f_i})\inverse \bar{Y}_{n_i}$ from \eqref{eq:GPGSmean} and $(\bar{K}_{a,a}|\bar{K}_{f_i,f_i})\inverse$ in \eqref{eq:GPGSvariance}.

\subsection{Real-time Simulation}
\label{sec:realtime_simulation}

For each step (frame) in the simulation, we will receive the status of the quadrotor $\prescript{\mathcal{W}}{}{x}_q=\stack(r_{q,1}, r_{q,2},r_{q,3},r_{q,4},\psi_q,\theta_q,\phi_q,v_{q,x},v_{q,y},v_{q,z})$ in the world frame $\mathcal{W}$. It is important to notice that during real-time simulation, airspeed is represented as a three-dimensional vector $v=\stack(v_{q,x},v_{q,y},v_{q,z})$. This airspeed is the sum of the quadrotor's velocity and the wind. Therefore, before performing the GP regression, we must ensure that the relationship between the quadrotor's body frame and its airspeed vector is consistent with that used in the training dataset.

To achieve this, we try to align the airspeed direction with the $+x$ axis in a pseudo-intermediate frame $\mathcal{H}$ by computing a rotation matrix that does so. We employ the \emph{Householder rotation} \cite{yang2021multi} to efficiently construct the alignment rotation matrix $\Wframe{R}_{\mathcal{H}}$:

\begin{equation}\begin{aligned}
    \hat{v} = \frac{v}{\norm{v}},&\quad u=\frac{\hat{v}-e_x}{\norm{\hat{v}-e_x}},\\
    \Wframe{R}_{\mathcal{H}}&= I-2uu\transpose,
\end{aligned}
\end{equation}
where $e_x=\smallbmat{1\\0\\0}$.

Next, to obtain the Euler angle representation of the quadrotor in this intermediate coordinate frame $\mathcal{H}$, we remove the above rotation component from the full rotation between the world frame and the body frame $\prescript{\mathcal{W}}{}{R}_\mathcal{B}$, and back-propagate the remaining rotation to express the orientation of the quadrotor within the intermediate frame $\mathcal{H}$:\begin{equation}
    \begin{aligned}
        \prescript{\mathcal{W}}{}{R}_{\mathcal{B}} &= R_z(\psi_q) R_y(\theta_q) R_x(\phi_q),\\
        \prescript{\mathcal{H}}{}{R}_\mathcal{B}&=\prescript{\mathcal{W}}{}{R}_\mathcal{H}\transpose\prescript{\mathcal{W}}{}{R}_\mathcal{B},\\
         \mathrm{Euler}_{\mathcal{C}\!/\!\mathcal{R}}(\prescript{\mathcal{H}}{}{R}_\mathcal{B})&=(\psi_h,\theta_h,\phi_h),
    \end{aligned}
\end{equation}

where $\mathrm{Euler}_{\mathcal{C}\!/\!\mathcal{R}}(*)$ is the algorithm that converts a rotation matrix to the corresponding Euler angles.

At this stage, we obtain the corrected state representation that can be directly used for Gaussian Process regression during real-time prediction:
\begin{equation}
    \prescript{\mathcal{H}}{}{x}_{q}=\stack(r_{q,1}, r_{q,2},r_{q,3},r_{q,4},\psi_h,\theta_h,\phi_h,\norm{\mathbf{v}}).
\end{equation}

We first compute the dynamics from the simplified model $\mathcal{M}$ as $y_\mathcal{M}=\stack(\prescript{\mathcal{W}}{}{F}_\mathcal{M}, \prescript{\mathcal{W}}{}{\tau}_\mathcal{M})$.

For the GP regression, we first check which bin the current query point $\prescript{\mathcal{H}}{}{x}_{q}$ falls into. Assuming that the query point exists in $b_i$, we compute the mean $\prescript{\mathcal{H}}{}{\mu_{q|a,b_i}}$ as shown in \eqref{eq:GPGSmean} and the variance $K_{q|a,b_i}$ as \eqref{eq:GPGSvariance} based on $\prescript{\mathcal{H}}{}{x}_{q}$.

The final prediction will need to be transferred back to the world frame and reintroduce $y_\mathcal{M}$ and leave the noise mean prediction from the GP model unaltered:
\begin{equation}
    \prescript{\mathcal{W}}{}{\mu_{q|a,b_i}} = \begin{bmatrix}
        \prescript{\mathcal{W}}{}{R}_\mathcal{H}\prescript{\mathcal{H}}{}{(\mu_{q|a,b_i})}_{[1:3]}+\prescript{\mathcal{W}}{}{F}_\mathcal{M}\\
        \prescript{\mathcal{W}}{}{R}_\mathcal{H}\prescript{\mathcal{H}}{}{(\mu_{q|a,b_i})}_{[4:6]}+\prescript{\mathcal{W}}{}{\tau}_\mathcal{M}\\
        (\mu_{q|a,b_i})_{[7:9]}
    \end{bmatrix}.
\end{equation}

\begin{figure}[h]
  \newcommand{\picwidth}{\linewidth}
  \centering
  \subfloat[Accuracy for force $F_x$]{%
    \includegraphics[width=0.8\picwidth]{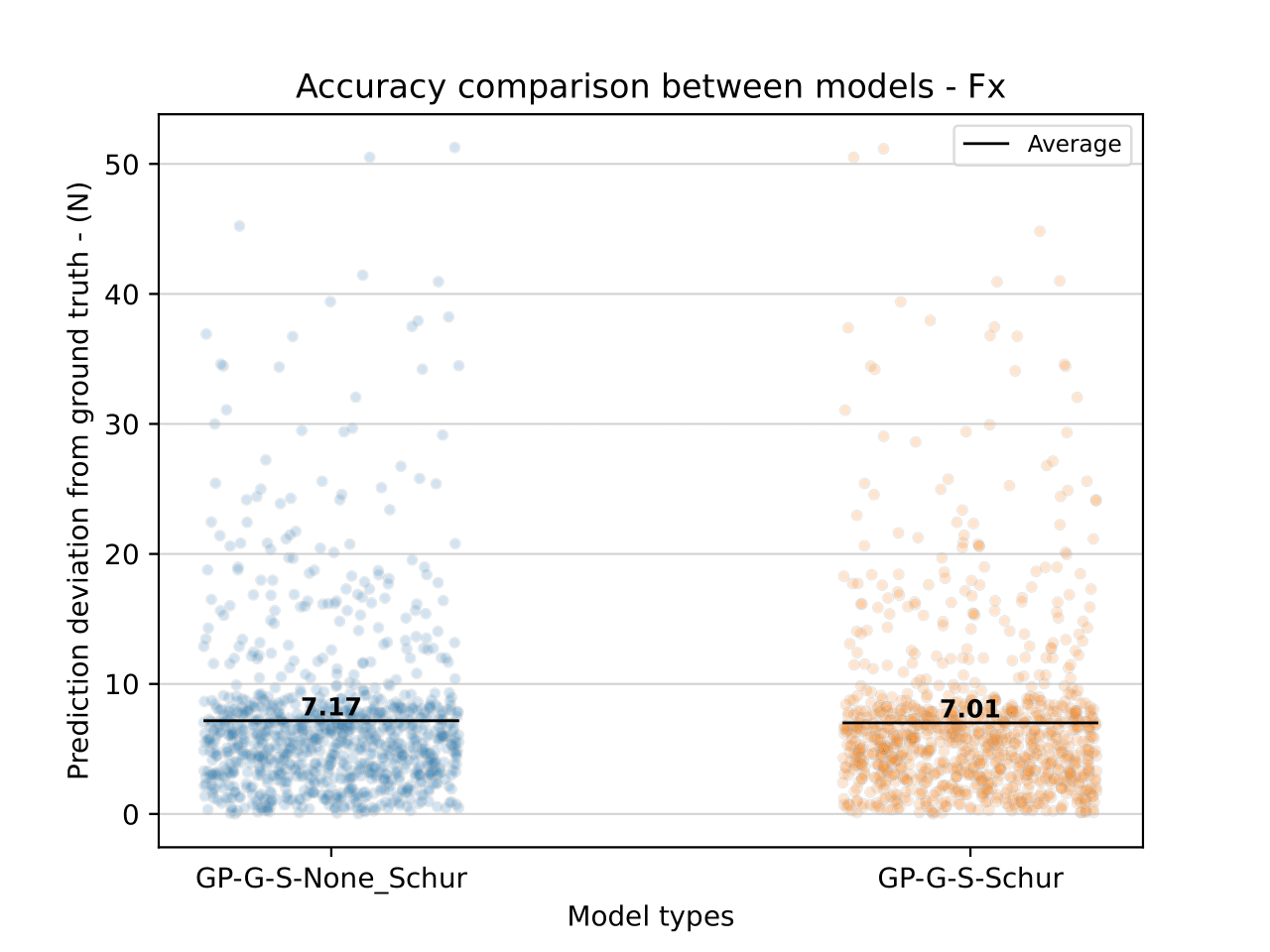}%
    \label{fig:schur_compare_Fx}%
  }

  \subfloat[Accuracy for torque $\tau_x$]{%
    \includegraphics[width=0.8\picwidth]{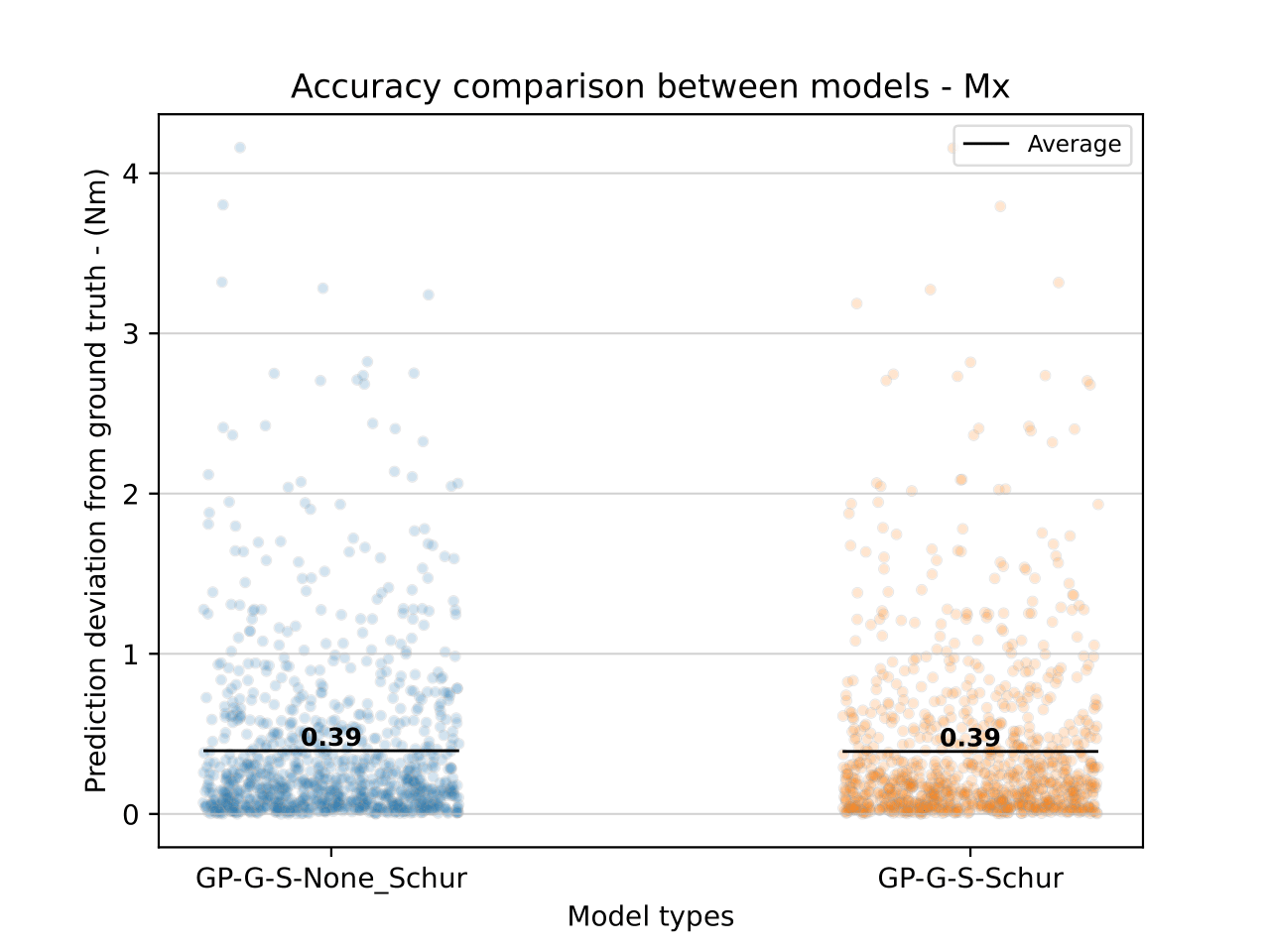}%
    \label{fig:schur_compare_Mx}%
  }

  \subfloat[Accuracy for noise $L_1$]{%
      \includegraphics[width=0.8\picwidth]{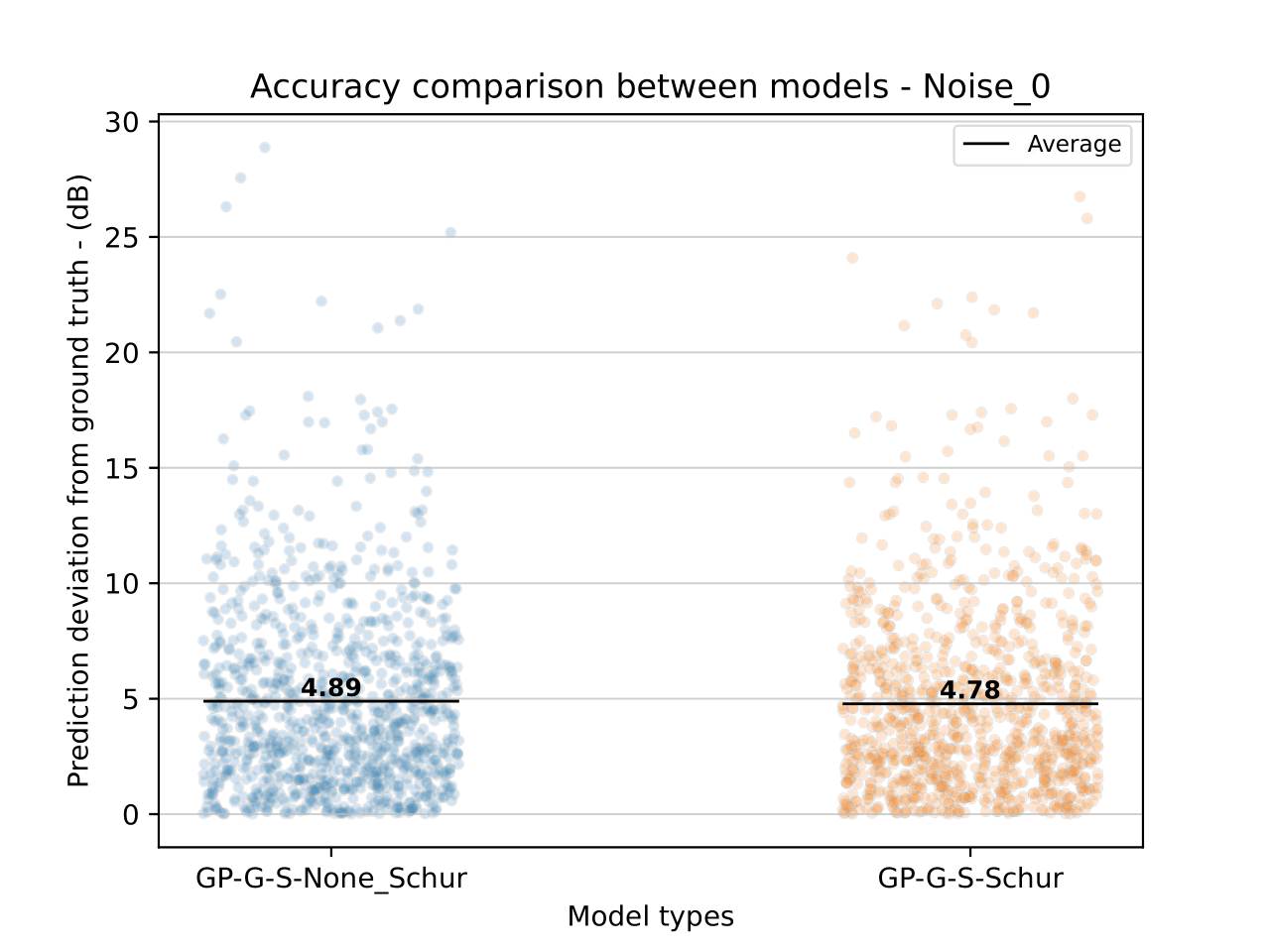}%
      \label{fig:schur_compare_Noise}%
  }

  \caption{Schur complement comparison}
  \label{fig:schur_compare}
\end{figure}

\section{Results}

In this section, we present a performance comparison between our proposed model and several GP baselines. The comparison includes:
\begin{itemize}
\item \textbf{GP-G-S-Schur}: our approach, a partitioned GP with gradient conditioning and Schur complement,
\end{itemize}
together with several baselines for comparison:
\begin{itemize}
\item \textbf{GP-G-S-None Schur}: a partitioned GP with gradient conditioning,
\item \textbf{GP}: a standard GP without input partitioning,
\item \textbf{GP-G}: a GP with gradient conditioning but without input partitioning,
\item \textbf{GP-S}: a partitioned GP without gradient information.
\end{itemize}

We employed the CHARM simulation framework described earlier, using NASA’s SUI Endurance quadrotor as the test platform. A total of 5000 data samples with gradient information were collected for training. To ensure a fair comparison across models, we provided additional data to the models that cannot incorporate gradient information, since each gradient-augmented sample inherently contains more information. Specifically, because gradient data were collected along seven input dimensions, we supplied the non-gradient models with the original dataset without numerical gradient processing (a total of 40000 data points) to compensate for this information difference.

To show the additional information gain obtained by incorporating gradient information, we consider two versions of the \textbf{GP-S}:\begin{itemize}
    \item \textbf{GP-S-8X} trained with 40000 data points,
    \item \textbf{GP-S-1X} trained with 5000 data points.
\end{itemize}

An additional 1000 data points were collected as a test set. We compare the distributions of absolute errors between model predictions and simulated ground-truth values. Furthermore, since our goal is real-time dynamic prediction, we also evaluate and compare the computational cost of each model during inference.

\subsection{Schur Complement Effectiveness}

\Cref{fig:schur_compare} compares the prediction accuracy of the models \textbf{GP-G-S} with and without \emph{Schur complement correction} described in \Cref{sec:SchurPredict}.

It can be seen that the \emph{Schur complement} process improves the prediction accuracy across most output dimensions. Even in dimensions where the improvement is less pronounced, the prediction accuracy does not degrade below that of the original model. This demonstrates that the \emph{Schur complement} correction effectively enhances model fidelity without introducing instability.

\subsection{Accuracy}

\begin{figure}[h]
  \newcommand{\picwidth}{\linewidth}
  \centering
  \subfloat[Accuracy for force $F_x$]{%
    \includegraphics[width=0.8\picwidth]{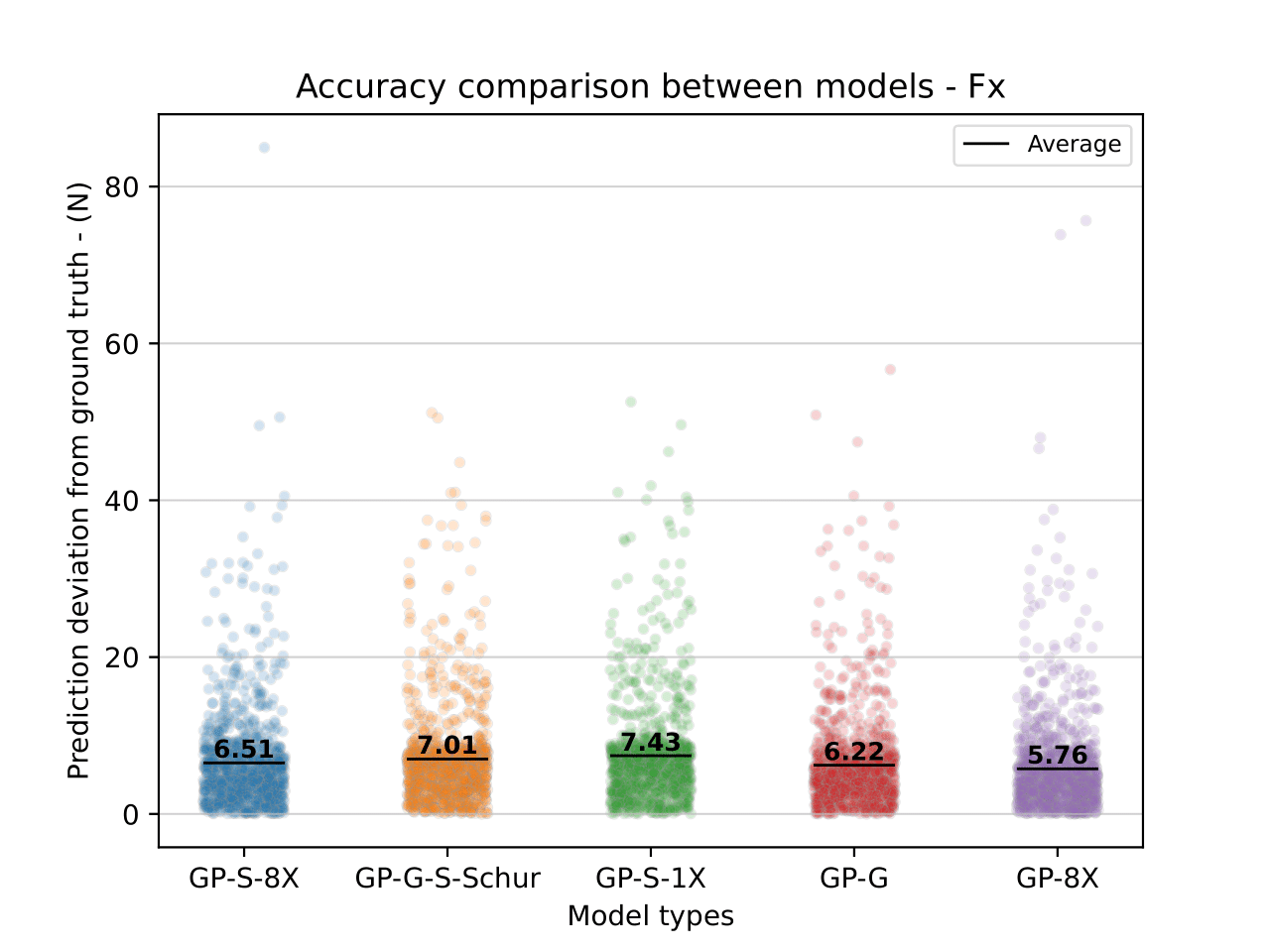}%
    \label{fig:model_compare_Fx}%
  }

  \subfloat[Accuracy for torque $\tau_x$]{%
    \includegraphics[width=0.8\picwidth]{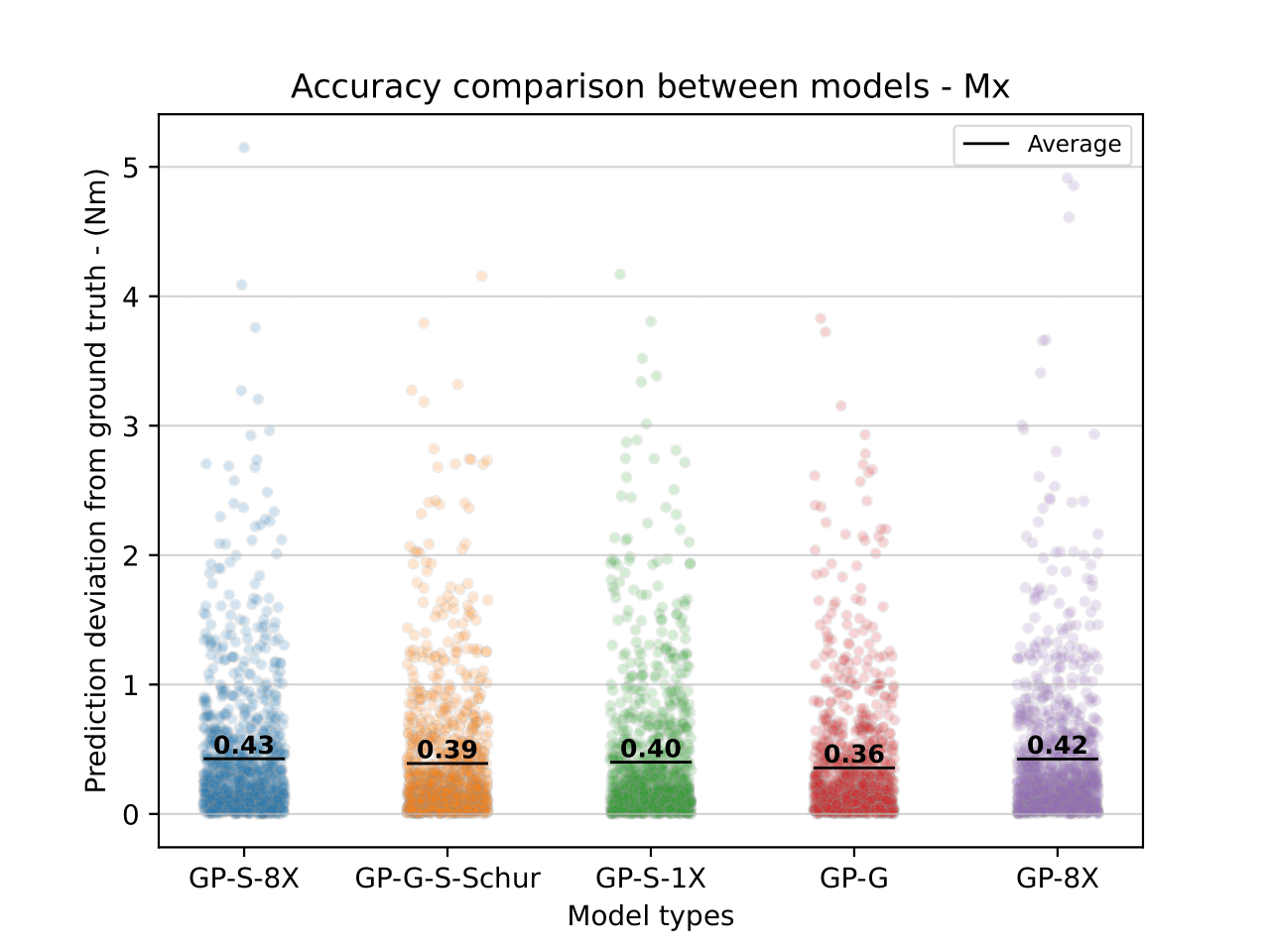}%
    \label{fig:model_compare_Mx}%
  }

  \subfloat[Accuracy for noise $L_1$]{%
      \includegraphics[width=0.8\picwidth]{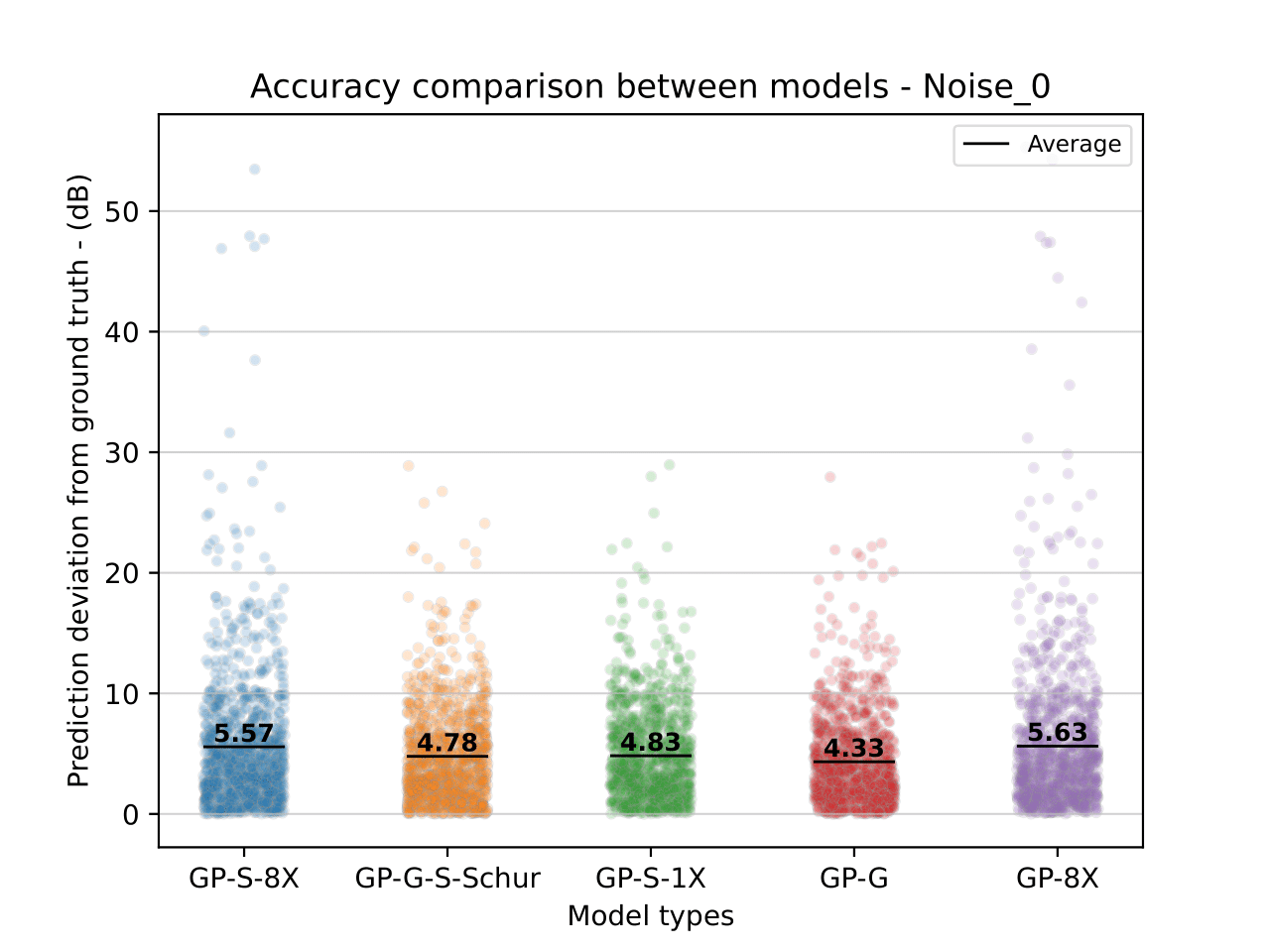}%
      \label{fig:model_compare_Noise}%
  }

  \caption{Model prediction comparison}
  \label{fig:model_compare_all}
\end{figure}

\Cref{fig:model_compare_all} illustrates the comparative prediction accuracy of four models in terms of force, torque, and noise, respectively.

It can be observed that, compared to \textbf{GP-G}, our method exhibits a slight loss in accuracy. However, compared with the \textbf{GP-S-1X} (and in some case even \textbf{GP-S-8X}), our approach achieves a notable improvement. This result indicates that incorporating gradient information and the Schur complement provides additional structural knowledge, leading to more accurate predictions.

At the same time, it can be observed that all models perform poorly at predicting noise. We attribute this to the absence of a simple baseline model for noise, unlike the force and torque predictions, which do have such a model. Given the limited amount of available data and the absence of a simple baseline model for noise, the GP models appear to reach their representational limits for this component.
\begin{figure}[h]
    \centering
    \includegraphics[page=2,width=\linewidth]{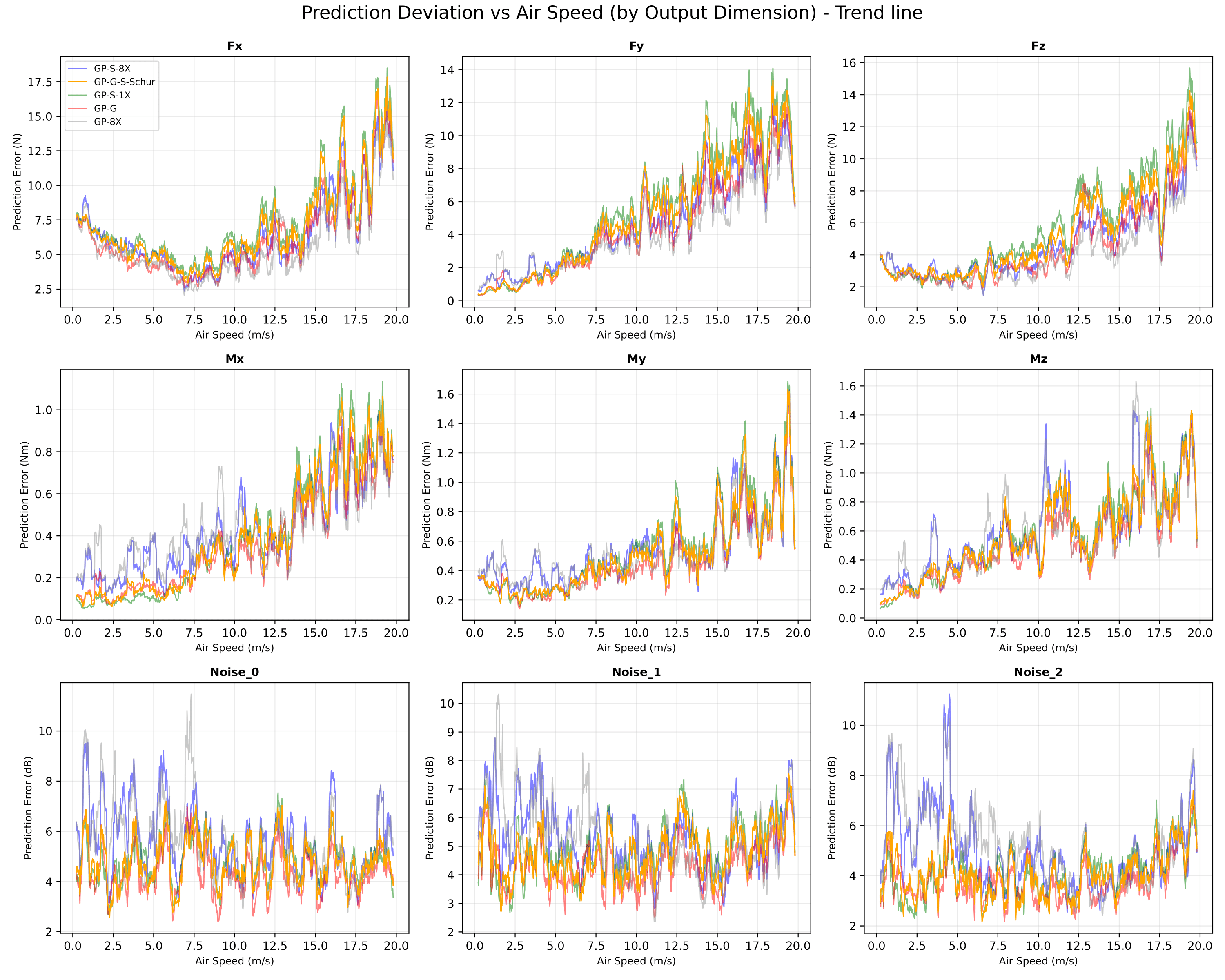}
    \caption{Prediction deviation (trend line) vs airspeed}
    \label{fig:air_vs_std}
\end{figure}

We further examined the prediction deviation across different dimensions of the test dataset. It was observed that the prediction accuracy for both force and torque deteriorates noticeably as the airspeed increases, whereas the noise prediction remains largely unaffected by changes in airspeed, as shown in \Cref{fig:air_vs_std}. We attribute this to the thickness noise component that is purely determined by blade geometry and rotation rate, independent of aerodynamic loading. However, since our dataset does not include the derivatives of the outputs with respect to airspeed, and the simplified model lacks explicit aerodynamic computations, the predictive performance for these two components decreases with higher airspeed.

\subsection{Computation time}

Our target is a simulation system capable of running at \unit[$30$]{Hz} in real time on commodity hardware (our simulation used a PC with an AMD 5800X CPU and 64 GB of DDR4 RAM, without GPU acceleration; multithreaded for matrix inversion, single-threaded for the rest). This requirement implies that each prediction of GP must be completed within $33$ milliseconds while maintaining acceptable accuracy.
\begin{table}[h!]
\centering % Use \centering instead of the center environment
\caption{Computational Time Comparison}
    \label{tab:model_compare_time}\begin{tabular}{l c c c} % 'l' for left-aligning the first column
    \toprule % Top rule
    \textbf{Model} & \textbf{Training (s)} & \textbf{Predicting (ms)} & Predicting\\
    \midrule % Middle rule
    GP-G-S-Schur & 1340.82 & 18.48 & 5.0\%\\
    GP-S-8X & 1422.15 & 35.27 & 9.4\%\\
    GP-G & 449.17 & 370.29 & 99.6\%\\
    GP & 453.59 & 371.72 & 100\%\\
    \bottomrule % Bottom rule
\end{tabular}
\end{table}

Table \ref{tab:model_compare_time} presents the median computation times for training and predictions in different models. These models are trained with 5000 (with gradient information) or 40000 (without gradient information) data points; all partitioned models are partitioned into 10 subregions.

Even compared to the partitioned standard GP model, our approach achieves nearly a $50\%$ improvement in computational speed with better accuracy. We attribute this advantage to the fact that each near group in our method contains only $1/8$ of the total data points, significantly reducing the number of kernel evaluations required to compute correlations between the query point and nearby samples.

With a median computation time of 18.48 milliseconds per prediction, our model enables real-time operation at a \unit[$30$]{Hz} update rate while leaving sufficient computational headroom for other system processes.

For training these models, a system with at least 64 GB of RAM is required for a dataset of 5000 data points with gradient information or 40000 without gradient information. In deployment, however, the partitioned GP with gradient information requires only 20 GB of RAM to load all models, compared to 25 GB for a single standard GP.

\begin{figure}[p]
  \newcommand{\picwidth}{\linewidth}
  \centering
  \subfloat[Simplified quadrotor dynamics]{%
      \includegraphics[width=\picwidth]{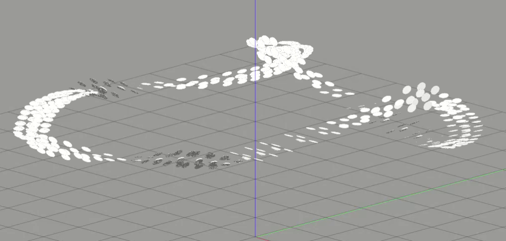}%
      \label{fig:Simple_sim}%
  }
  \hfill
  \subfloat[GP-based surrogate dynamics]{%
      \includegraphics[width=\picwidth]{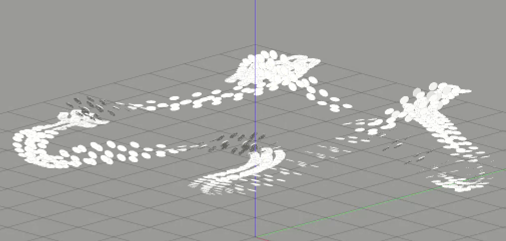}%
      \label{fig:GP_sim}%
  }
  \hfill
  \subfloat[Flight path comparison between 2 waypoints project on $yz$ plane]{%
      \includegraphics[page=1,width=\picwidth]{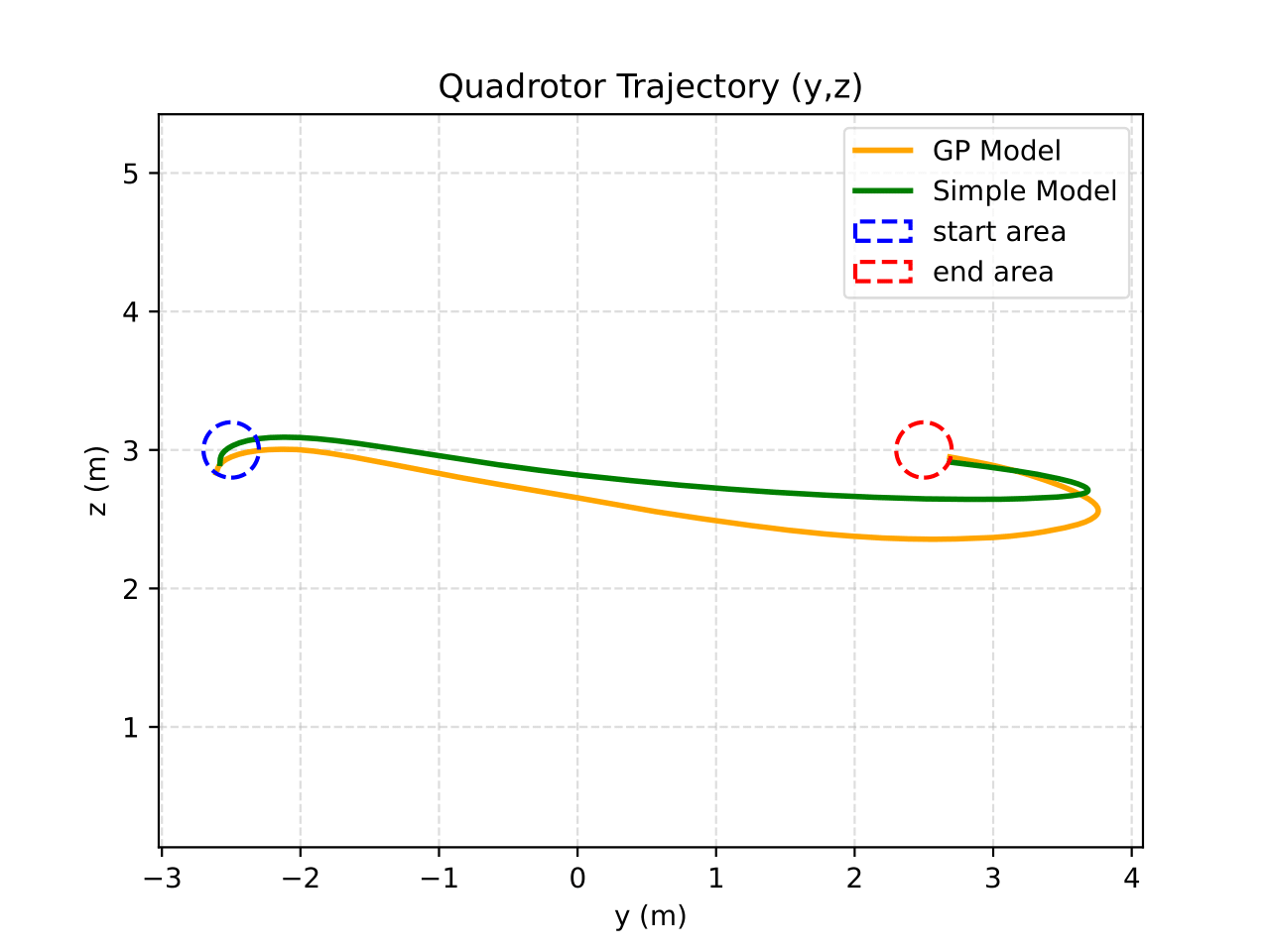}%
      \label{fig:Flight_compare_yz}%
  }
  % \hfill
  % \subfloat[Flight path comparison between 2 waypoints project on $yx$ plane]{%
  %     \includegraphics[width=\picwidth]{figures/model_compare_traj_yx-1.png}%
  %     \label{fig:Flight_compare_yx}%
  % }
  \hfill
  \subfloat[Average rotor RPM between two waypoints]{%
      \includegraphics[width=\picwidth]{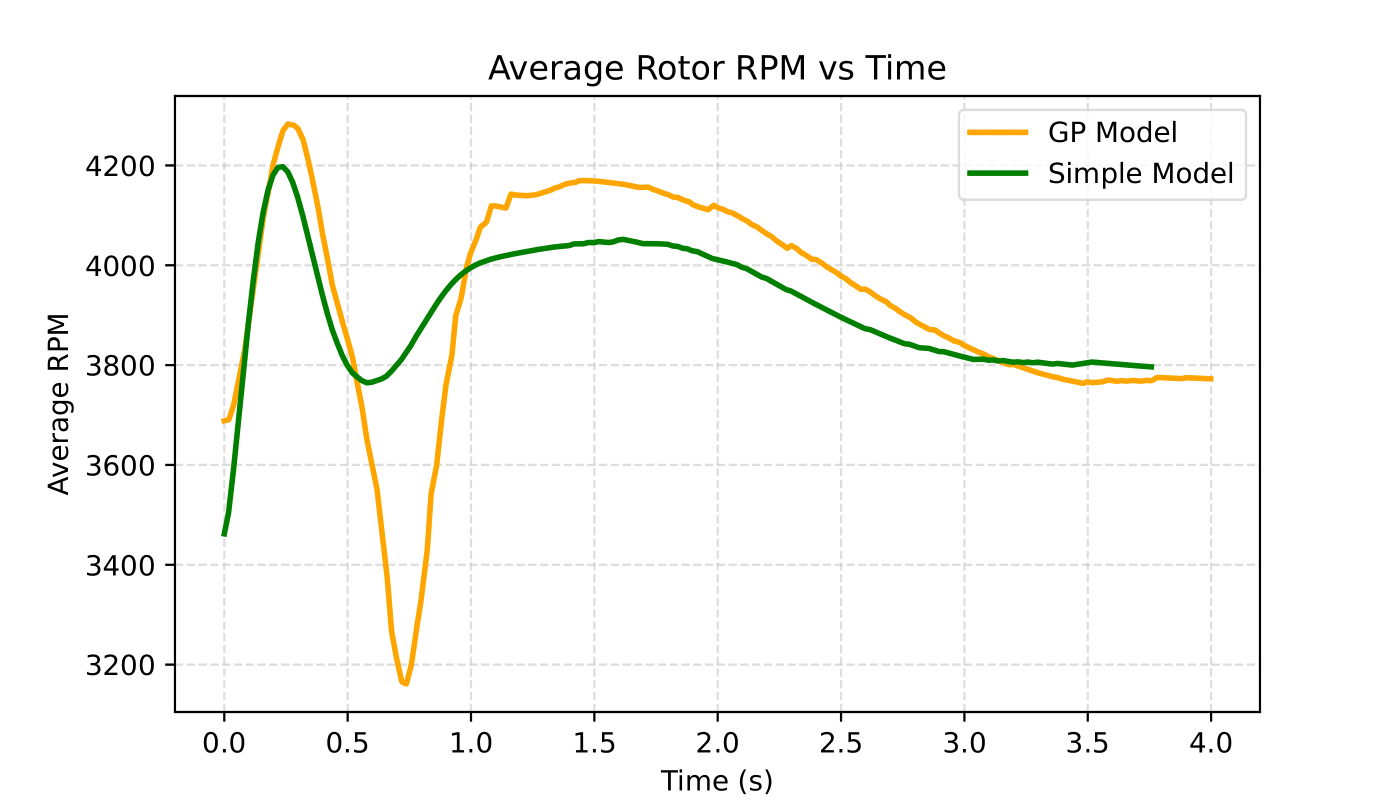}%
      \label{fig:Flight_compare_rpm}%
  }
  \caption{Comparison of quadrotor dynamics simulation in Gazebo}
  \label{fig:quadrotor_dynamics_comparison}
\end{figure}

\subsection{Simulation Environment}

\Cref{fig:quadrotor_dynamics_comparison} illustrates quadrotor simulations based on the simplified dynamics model and our GP–based dynamics model, respectively. Both simulations employ the same geometric control algorithm \cite{mellinger2011minimum}, which is designed for the simplified dynamics model (\Cref{sec:SimpleModel}). As shown in \Cref{fig:Flight_compare_yz}, the controller performs well in the simplified model, enabling the quadrotor to traverse between waypoints smoothly. In contrast, in the GP-based simulation, the quadrotor experiences noticeable altitude loss and overshoot when tilting toward a waypoint due to aerodynamic drag, and takes longer to precisely reach each waypoint. The average rpm of each rotor also experiences larger fluctuations and always requires more power to get the quadrotor to the desired waypoint (\Cref{fig:Flight_compare_rpm}). This observation demonstrates that our GP-based dynamics simulation achieves substantially greater physical fidelity than the simplified model, providing a more realistic environment for the development of quadrotor controllers.

\section{Conclusion}

We have developed a partitioned Gaussian Process framework with gradient conditioning for real-time prediction of quadrotor dynamics. By combining gradient information, Schur complement approximations, and state-space partitioning, our method significantly reduces computational complexity while preserving high predictive accuracy. The inclusion of gradient information provides additional structural insight into the aerodynamic behavior, enabling the model to outperform standard GPs with comparable or reduced datasets. Offline precomputation of the covariance structures for each partitioned region further enhances runtime efficiency, allowing real-time updates at \unit[$30$]{Hz} with average prediction times under \unit[20]{ms}. Although the framework achieves robust force and torque predictions, performance on noise estimation remains limited due to insufficient data and the absence of a simplified noise model. Future work will extend this approach by incorporating a simplified noise model to improve acoustic noise prediction accuracy and by applying a smaller partition interval in high airspeed conditions to generate smaller bins for better prediction of the fast-changing aerodynamic characteristics.

\bibliographystyle{biblio/ieee}

\bibliography{biblio/IEEEfull,biblio/IEEEConfFull,biblio/OtherFull,% Do not insert spaces in this command, otherwise it will not work.
  biblio/tron,%
  biblio/formationControl,%
  biblio/websites,%
  biblio/gaussianProcess,%
  biblio/quadrotorApplications,biblio/quadrotorControl,%
  biblio/charm%
}

\end{document}

%% file: preamble/common.tex
\input{preamble/commonNoTikz}
\input{preamble/graphicsTikz}
\input{preamble/robotics}

%% file: preamble/commonNoTikz.tex
\input{preamble/fixes}
\input{preamble/graphics}
\input{preamble/pagination}
\input{preamble/math}
\input{preamble/operators}
\input{preamble/notation}
\input{preamble/units}
\input{preamble/markupAndCommenting}

\input{preamble/formattingCode}

\input{preamble/utilities}

%% file: preamble/fixes.tex
\usepackage{fix-cm}
\usepackage{etex}

\usepackage{dblfloatfix}

\usepackage{nag}

%% file: preamble/graphics.tex
\makeatletter
\@ifpackageloaded{xcolor}{}{%
\usepackage[table,x11names,dvipsnames,svgnames]{xcolor}%
}
\makeatother

\usepackage{colortbl}

\usepackage{graphicx}
\graphicspath{{./figures/}}

\usepackage{wrapfig}

\input{preamble/graphicsColors}

%% file: preamble/graphicsColors.tex
\definecolorset{RGB}{lyft}{}{Red,194,39,36;Sunset,202,53,33;Orange,205,68,20;Amber,200,117,42;Yellow,242,169,52;Citron,186,188,44;Lime,112,159,33;Green,56,139,31;Mint,45,118,56;Teal,52,133,135;Cyan,60,132,202;Blue,55,94,248;Indigo,64,13,247;Purple,115,42,248;Pink,176,25,145;Rose,176,32,75}

\definecolorset{HTML}{h}{}{grapefruit,ED5565;bittersweet,FC6E51;sunflower,FFCE54;grass,A0D468;mint,48CFAD;aqua,4FC1E9;bluejeans,5D9CEC;lavender,AC92EC;pinkrose,EC87C0;lightgray,F5F7FA;mediumgray,CCD1D9;darkgray,656D78}

\definecolorset{RGB}{p}{}{ivory,221,221,221;navy,046,037,133;pine,051,117,056;aqua,093,168,153;azure,148,203,236;sand,220,205,125;peach,194,106,119;carnation,159,074,150;mulberry,126,041,084}

\definecolorset{RGB}{o}{}{black,000,000,000;mint,000,158,115;jeans,000,114,178;azure,086,180,233;banana,240,228,066;orange,230,159,000;ember,213,094,000;amaranth,204,121,167}

%% file: preamble/pagination.tex
\usepackage{cite}

\usepackage{microtype}

\usepackage[american]{babel}

\usepackage{array}
\usepackage{multirow}
\usepackage{booktabs}
\usepackage{makecell} %introduces \thead and \makecell

\ifcsname labelindent\endcsname
\let\labelindent\relax
\fi
\usepackage[inline]{enumitem}

\usepackage{subfig}

\newenvironment{lenumerate}[2][]
{\begin{enumerate}[label=(#2\arabic*),leftmargin=0.2in,itemindent=0.15in,#1]}
{\end{enumerate}}

\setlist*[enumerate,1]{label={\itshape\arabic*)}}

\makeatletter
\newcommand{\paragraphswithstop}{%
\let\copyparagraph\paragraph%
\renewcommand\paragraph[1]{\copyparagraph{##1.}}%
}
\makeatother

\usepackage[framemethod=tikz]{mdframed}

\makeatletter
\def\namedlabel#1#2{\begingroup
  #2%
  \def\@currentlabel{#2}%
  \phantomsection\label{#1}\endgroup
}
\makeatother

\makeatletter
\def\namedlabelphantom#1#2{\begingroup
  \def\@currentlabel{#2}%
  \phantomsection\label{#1}\endgroup
}
\makeatother

\newcommand{\parunskip}{\bgroup\unskip\parfillskip=0pt \par\egroup}

%% file: preamble/operators.tex
\newcommand{\real}[1]{\mathbb{R}^{#1}{}}

\newcommand{\smallbmat}[1]{\left[\begin{smallmatrix}#1\end{smallmatrix}\right]}

\newcommand{\transpose}{^{\top}}

\newcommand{\inverse}{^{-1}}

\DeclarePairedDelimiter{\norm}{\lVert}{\rVert}

\DeclareMathOperator{\stack}{stack}

\newcommand{\union}{\cup}

%% file: preamble/notation.tex
\providecommand{\cB}{\mathcal{B}}

\providecommand{\cM}{\mathcal{M}}
\providecommand{\cN}{\mathcal{N}}

\providecommand{\cW}{\mathcal{W}}

%% file: preamble/units.tex
\usepackage{units}

%% file: preamble/markupAndCommenting.tex
  \newcommand{\newcolorlabel}[2]{%
  \expandafter\newcommand\csname #1\endcsname[1]{%
    \tikz[baseline]{\node[text=white,fill=#2,anchor=base,text height=1.3ex,text depth=0.1ex,font=\sffamily\bfseries]{##1}}}%
}

\newcommand{\newcommenter}[2]{%
  \expandafter\newcommand\csname #1\endcsname[1]{%
    \fcolorbox{#2}{#2}{\color{white}\textsf{\textbf{#1}}}
    {\color{#2}##1}}%
  \expandafter\newcommand\csname #1p\endcsname[1]{%
    \pdfcomment[color=#2,voffset=1Em]{#1:##1}}%
  \expandafter\newcommand\csname at#1\endcsname{%
    \fcolorbox{#2}{#2}{\color{white}\textsf{\textbf{@#1}}}
    {\color{#2}}}%
  \expandafter\newcommand\csname #1cite\endcsname[1]{%
    \csname #1\endcsname {[##1]}
  }%
  \expandafter\newcommand\csname #1ref\endcsname[1]{%
    \csname #1\endcsname {$\blacktriangleright$##1}
  }%
  \expandafter\newcommand\csname #1hl\endcsname[2]{%
    \colorbox{#2}{\color{white}\textsf{\textbf{#1}}}\sethlcolor{Azure2}\hl{##2}~%
    \expandafter\ifx\csname commentarrow\endcsname\relax$\leftarrow$\else \commentarrow[#2]\fi~%
    {\color{#2}##1}}%
  \expandafter\newcommand\csname #1hlp\endcsname[2]{%
    \sethlcolor{#2!50!white}\hl{##2}%
    \pdfmargincomment[color=#2,voffset=1Em,icon=Note,open=true,author=#1]{#1:##1}}%
  \expandafter\newcommand\csname #1st\endcsname[2]{%
    \colorbox{#2}{\color{white}\textsf{\textbf{#1}}}\sout{##2}~%
    \expandafter\ifx\csname commentarrow\endcsname\relax$\leftarrow$\else \commentarrow[#2]\fi~%
    {\color{#2}##1}}%
}
\newcommenter{TODO}{DodgerBlue1}
\newcommenter{rtron}{OliveDrab2}

\usepackage{comment}

\usepackage{pdfcomment}

\usepackage{soul}

\usepackage[normalem]{ulem}

\usepackage{csquotes}

%% file: preamble/formattingCode.tex
\usepackage{listings}

%% file: preamble/utilities.tex
\usepackage{suffix}

\usepackage{environ}

\makeatletter
\newsavebox{\boxifnotempty}
\newcommand{\displayifnotempty}[3]{\sbox\boxifnotempty{#2}\setbox0=\hbox{\usebox{\boxifnotempty}\unskip}%
  \ifdim\wd0=0pt
  \else
  #1\usebox{\boxifnotempty}#3%
  \fi%
}

\newcommand{\ifempty}[2]{\setbox0=\hbox{#1\unskip}%
  \ifdim\wd0=0pt%
  #2%
  \fi%
}

\newcommand{\ifnotempty}[2]{\setbox0=\hbox{#1\unskip}%
  \ifdim\wd0>0pt%
  #2%
  \fi%
}
\makeatother

\newcommand{\switchifempty}[3]{\sbox\boxifnotempty{#1}\setbox0=\hbox{\usebox{\boxifnotempty}\unskip}%
  \ifdim\wd0=0pt{}%
  #2%
  \else{}%
  #3%
  \usebox{\boxifnotempty}%
  \fi%
}

\makeatletter
\@ifundefined{chapter}{\usepackage{algorithm}}{\usepackage[chapter]{algorithm}}
\makeatother
\usepackage{algorithmicx}
\usepackage{algpseudocode}
\makeatother%

\usepackage{scrlfile}

\makeatletter
\newcommand*\newstoreddef[1]{
  \BeforeClosingMainAux{%
    \immediate\write\@auxout{%
      \string\restoredef{#1}{\csname #1\endcsname}%
    }%
  }%
}
\newcommand*{\restoredef}[2]{% used at the aux file
  \expandafter\gdef\csname stored@#1\endcsname{#2}%
}
\newcommand*{\storeddef}[1]{
  \@ifundefined{stored@#1}{0}{\csname stored@#1\endcsname}%
}
\makeatother

\usepackage{pageslts}
\NewEnviron{tee}{\BODY\typeout{Marker Tee [start] ^^J \BODY ^^JMaker Tee [end]}}

\usepackage{cleveref}
\crefname{property}{property}{properties}
\Crefname{property}{Property}{Properties}
\crefname{assumption}{assumption}{assumptions}
\Crefname{assumption}{Assumption}{Assumptions}
\crefname{problem}{problem}{problems}
\Crefname{problem}{Problem}{Problems}
\crefname{fact}{fact}{facts}
\Crefname{fact}{Fact}{Facts}
\crefname{remark}{remark}{remarks}
\Crefname{remark}{Remark}{Remarks}
\usepackage{nameref}
\makeatletter
\@ifundefined{corollary}{}{
  \crefname{corollary}{corollary}{corollary}
  \Crefname{corollary}{Corollary}{Corollaries}
  \crefname{example}{example}{examples}
  \Crefname{example}{Example}{Examples}
  \crefname{remark}{remark}{remarks}
  \Crefname{remark}{Remark}{Remarks}
  \crefname{property}{property}{properties}
  \Crefname{property}{Property}{Properties}
  \crefname{assumption}{assumption}{assumptions}
  \Crefname{assumption}{Assumption}{Assumptions}
  \crefname{problem}{problem}{problems}
  \Crefname{problem}{Problem}{Problems}
  \crefname{fact}{fact}{facts}
  \Crefname{fact}{Fact}{Facts}
  \crefname{lemma}{lemma}{lemmas}
  \Crefname{lemma}{Lemma}{Lemmas}
  \crefname{proposition}{proposition}{propositions}
  \Crefname{proposition}{Proposition}{Propositions}
}

%% file: preamble/graphicsTikz.tex
\usepackage{tikz}
\usetikzlibrary{calc}
\usetikzlibrary{matrix}
\usetikzlibrary{chains,scopes}
\usetikzlibrary{shapes.geometric}
\usetikzlibrary{arrows.meta}
\usetikzlibrary{decorations.markings}
\usetikzlibrary{decorations.pathreplacing}
\usetikzlibrary{backgrounds}

\tikzset{
  dim above/.style={to path={\pgfextra{
        \pgfinterruptpath
        \draw[>=latex,|->|] let
        \p1=($(\tikztostart)!1.5em!90:(\tikztotarget)$),
        \p2=($(\tikztotarget)!1.5em!-90:(\tikztostart)$)
        in(\p1) -- (\p2) node[pos=.5,sloped,above]{#1};
        \endpgfinterruptpath
      }
    }
  },
  dim double above/.style={to path={\pgfextra{
        \pgfinterruptpath
        \draw[>=latex,|->|] let
        \p1=($(\tikztostart)!3em!90:(\tikztotarget)$),
        \p2=($(\tikztotarget)!3em!-90:(\tikztostart)$)
        in(\p1) -- (\p2) node[pos=.5,sloped,above]{#1};
        \endpgfinterruptpath
      }
    }
  },
  dim below/.style={to path={\pgfextra{
        \pgfinterruptpath
        \draw[>=latex,|->|] let
        \p1=($(\tikztostart)!-1em!-90:(\tikztotarget)$),
        \p2=($(\tikztotarget)!-1em!90:(\tikztostart)$)
        in (\p1) -- (\p2) node[pos=.5,sloped,below]{#1};
        \endpgfinterruptpath
      }
    }
  },
}

\tikzset{
    right angle quadrant/.code={
        \pgfmathsetmacro\quadranta{{1,1,-1,-1}[#1-1]}     % Arrays for selecting quadrant
        \pgfmathsetmacro\quadrantb{{1,-1,-1,1}[#1-1]}},
    right angle quadrant=1, % Make sure it is set, even if not called explicitly
    right angle length/.code={\def\rightanglelength{#1}},   % Length of symbol
    right angle length=2ex, % Make sure it is set...
    right angle symbol/.style n args={3}{
        insert path={
            let \p0 = ($(#1)!(#3)!(#2)$) in     % Intersection
                let \p1 = ($(\p0)!\quadranta*\rightanglelength!(#3)$), % Point on base line
                \p2 = ($(\p0)!\quadrantb*\rightanglelength!(#2)$) in % Point on perpendicular line
                let \p3 = ($(\p1)+(\p2)-(\p0)$) in  % Corner point of symbol
            (\p1) -- (\p3) -- (\p2)
        }
    }
}

\newcommand{\pgfextractangle}[3]{%
    \pgfmathanglebetweenpoints{\pgfpointanchor{#2}{center}}
                              {\pgfpointanchor{#3}{center}}
    \global\let#1\pgfmathresult
}

\usetikzlibrary{shapes.arrows}
\newcommand{\commentarrow}[1][Azure4]{\tikz[baseline=-3pt]{\node[shape border uses incircle, fill=#1,rotate=180,single arrow, inner sep=1pt, minimum size=6pt, single arrow head extend=2pt]{};}}

%% file: preamble/robotics.tex
\input{preamble/roboticsGraphics}

\input{preamble/roboticsTerms}

%% file: preamble/roboticsGraphics.tex
\tikzset{ax/.style={-latex,line width=2pt}}

\tikzset{camera/.style={fill=Sienna1,fill opacity=0.5},%
image plane/.style={draw=RoyalBlue3,line width=2pt}}

%% file: preamble/roboticsTerms.tex
\newcommand{\Wframe}{\prescript{\cW}{}}

\newcommand{\Bframe}[1][]{\prescript{\cB_{#1}}{}}